%% file: kyamashita_pose_camera_ready.tex
\documentclass[runningheads]{llncs}

\usepackage{eccv}

\usepackage{eccvabbrv}

\usepackage{graphicx}
\usepackage{booktabs}

\usepackage[accsupp]{axessibility}  

\usepackage{colortbl}
\usepackage{wrapfig}
\usepackage{algorithmic}
\usepackage[ruled]{algorithm2e}
\usepackage {afterpage}

\usepackage{hyperref}

\usepackage{orcidlink}

\input{vincents_commands}
\addeditor{vincent}{VL}{0.0, 0.5, 0.0}
\addeditor{kohei}{KY}{0.0, 0.4, 0.6}
\addeditor{Ko}{KN}{1,0,1}
\showeditstrue

\newcommand{\refl}{\omega_\mathrm{r}}

\newcommand{\transp}{\mathrm{T}}
\textvars{Norm}
\newcommand{\IM}{\mathcal{I}}
\newcommand{\NM}{\mathcal{N}}
\newcommand{\RM}{\mathcal{R}}

\begin{document}

\title{Correspondences of the Third Kind: \\
    Camera Pose Estimation from Object Reflection} 

\titlerunning{Camera Pose Estimation from Object Reflection}

\author{Kohei Yamashita\inst{1}\orcidlink{0000-0002-5086-9906} \and
Vincent Lepetit\inst{2}\orcidlink{0000-0001-9985-4433} \and 
Ko Nishino\inst{1}\orcidlink{0000-0002-3534-3447}}

\authorrunning{K.~Yamashita et al.}

\institute{Graduate School of Informatics, Kyoto University, Kyoto, Japan \\
\email{kyamashita@vision.ist.i.kyoto-u.ac.jp kon@i.kyoto-u.ac.jp} 
\url{https://vision.ist.i.kyoto-u.ac.jp/} \and
LIGM, Ecole des Ponts, Univ Gustave Eiffel, CNRS, Marne-la-vall\'ee, France\\
\email{vincent.lepetit@enpc.fr} \\
}

\maketitle

\begin{abstract}
  Computer vision has long relied on two kinds of correspondences: pixel correspondences in images and 3D correspondences on object surfaces. Is there another kind, and if there is, what can they do for us? In this paper, we introduce correspondences of the third kind we call reflection correspondences and show that they can help estimate camera pose by just looking at objects without relying on the background. Reflection correspondences are point correspondences in the reflected world, \ie, the scene reflected by the object surface. The object geometry and reflectance alter the scene geometrically and radiometrically, respectively, causing incorrect pixel correspondences. Geometry recovered from each image is also hampered by distortions, namely generalized bas-relief ambiguity, leading to erroneous 3D correspondences. We show that reflection correspondences can resolve the ambiguities arising from these distortions. We introduce a neural correspondence estimator and a RANSAC algorithm that fully leverages all three kinds of correspondences for robust and accurate joint camera pose and object shape estimation just from the object appearance. The method expands the horizon of numerous downstream tasks, including camera pose estimation for appearance modeling~(\eg, NeRF) and motion estimation of reflective objects~(\eg, cars on the road), to name a few, as it relieves the requirement of overlapping background.
\end{abstract}

\section{Introduction}
\label{sec:intro}

\begin{wrapfigure}{r}{0.45\textwidth}
    \centering
    \includegraphics[width=\linewidth]{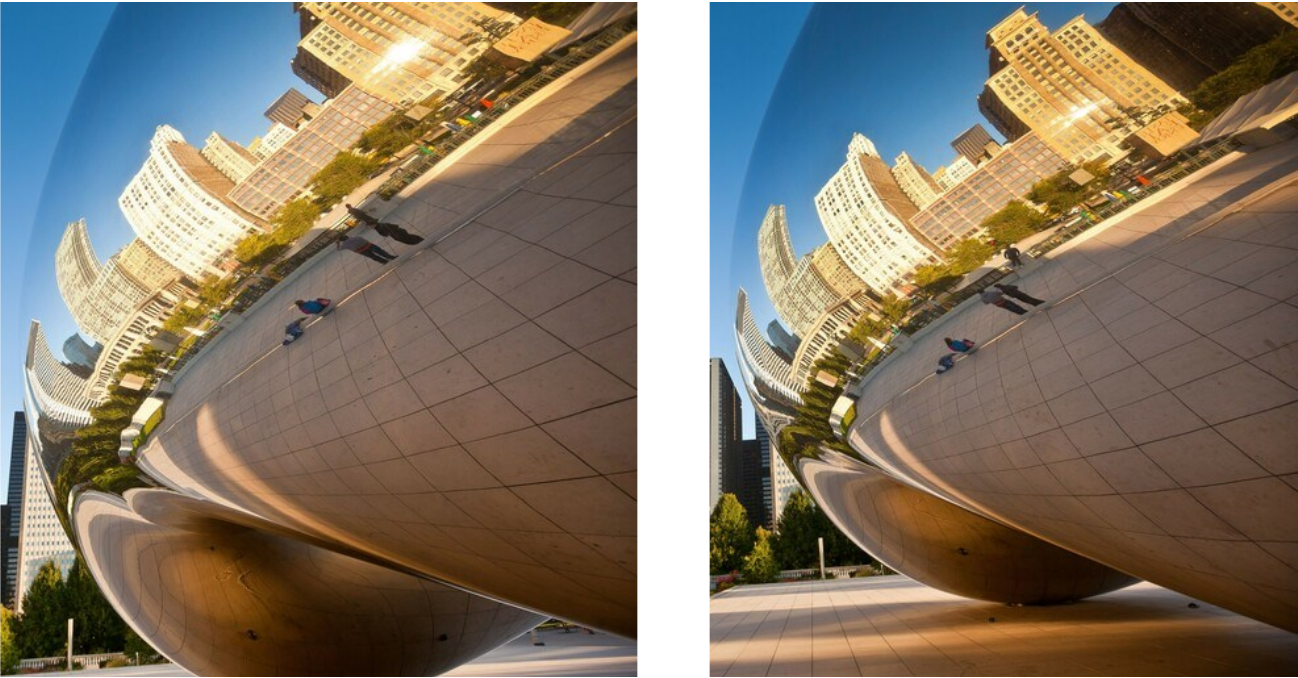}
    \caption{We humans can tell how the camera moved between the images, but computers have a hard time. Can we estimate camera pose and possibly object shape just from object appearance, despite the featureless appearance and non-overlapping background? (Photos by Richard Ellis/Alamy)}
    \label{fig:opening}
\end{wrapfigure}

Look at the two images in \cref{fig:opening}.
Even for this extreme case of a perfect mirror object, we---as humans---can understand how the camera moved (at least qualitatively) between the two images. This likely owes to our ability to disentangle the reflected surroundings from surface appearance.

How would a computer estimate the camera pose change between the two images? Structure-from-motion would fail in such conditions because they rely on pixel correspondences, \ie, matched projections of the same physical surface points in the images, which would be erroneous as the glossy reflection violates the color constancy constraint. 

Recent neural shape-from-shading methods can recover the object geometry as surface normals in each view even under complex natural illumination \cite{yamashita2023deepsharm}. As the illumination is unknown, the recovered surface normals, however, suffer from a fundamental ambiguity known as the generalized bas-relief ambiguity~\cite{belhumeur99basrelief} between light source and surface geometry. That is, a rotated illumination and sheared surface can conspire to generate the exact same object appearance. As such, 3D correspondences established between the two surface normal maps recovered independently from the two images would only tell us the camera pose in the distorted space.

In fact, we show that even if we use both pixel correspondences in the images and 3D correspondences on the recovered object surfaces, we cannot resolve this ambiguity for orthographic cameras. Even with a perspective camera, the object is often far enough that this limitation still holds. Even if we have more than two views, joint estimation of the camera poses and the object shape with the erroneous normal maps would be still challenging. The geometry estimate can easily fall into local minima whose surface normals (\ie, local surface geometry) are consistent with the current camera pose estimates. 

How then can the camera pose be estimated just from the object appearance? We take a hint from what we humans likely do: establish correspondences in the world reflected by the object surface. By leveraging the single-view reconstruction of surface normals, we can extract from each of the images a reflectance map. The reflectance map is the Gaussian sphere of the reflected radiance. It represents the surrounding environment modulated by surface reflectance as a spherical surface indexed by surface normals. We establish correspondences between the reflectance maps of the images. We refer to these as reflection correspondences. 

We show that these reflection correspondences by themselves are not sufficient to directly compute the camera pose. They, however, resolve the ambiguity remaining in the pixel and 3D correspondences due to the bas-relief ambiguity. This means that, by using all three types of correspondences in two or more images of an object of arbitrary reflectance, even without any texture and with strong specularity, we can compute the relative camera poses just from the object appearance. This liberates many applications from the seemingly benign yet practically extremely limiting requirement of overlapping static background or diffuse surface texture just to recover camera positions.

We formalize reflection correspondences and show how they should be combined with conventional correspondences (pixel correspondences and 3D correspondences) to obtain a quantitative estimate of the correct camera rotation. We introduce a RANSAC-based, two-step algorithm which first exploits conventional correspondences and then resolves the ambiguity with reflection correspondences. We also introduce a new neural feature extractor for establishing 3D and reflection correspondences robust to the inherent distortions primarily caused by the bas-relief ambiguity with effective data augmentation. Finally, we derive a joint estimation framework for accurate joint camera pose and geometry reconstruction which alternates between the interdependent two quantities.

Experimental results on synthetic and real images show that the reflection correspondences and the neural feature extractor as well as our iterative estimation framework are essential for accurate camera pose and geometry estimation.
We believe reflection correspondences can play an important role in applications beyond camera pose and shape recovery from object appearance, including camera calibration~\cite{Bhayani23} and object pose estimation~\cite{park-iccv19-pix2pose} when classical correspondences are not sufficient. 
All data and code can be found on our project web page.

\section{Related Work}

In this paper, we address camera pose estimation from a small number (2) of views of textureless, non-Lambertian (\eg, shiny) objects taken under unknown natural illumination, which remains challenging for existing methods especially when good initial estimates are not available.

\textbf{Structure-from-motion} methods recover camera poses for multiple images by detecting and leveraging pixel correspondences~\cite{SfM-survey, schoenberger2016colmap_sfm}. They typically detect such pixel correspondences by leveraging view-independent, salient features such as textures and jointly solve for camera poses and 3D locations of the surface points. Descriptors invariant to rotation and scale (\eg, SIFT~\cite{lowe04imagefeature}) and outlier correspondence detection with RANSAC~\cite{fischler81ransac} are often used for robust estimation. 
However, these methods are prone to fail on textureless, non-Lambertian objects (especially, shiny objects) as correspondences between surface points are extremely challenging to establish on these objects. Also, correspondences based on textures are usually sparse, and these methods require a large number (\eg, 50) of input images that have large visual overlaps for accurate estimation.

\textbf{Neural image synthesis} methods jointly estimate surface geometry and a surface light field (or its radiometric roots, \ie, reflectance and illumination) as neural representations with differentiable volumetric~\cite{mildenhall2020nerf, niemeyer2020dvr, zhang21nerfactor, wang2021neus, liu2023nero} or surface~\cite{yariv2020idr, zhanf2021physg} rendering. While most of these methods require multiple registered images as inputs, a few methods also handle non-registered multi-view images by optimizing the camera poses as additional parameters~\cite{yariv2020idr, boss2022samurai}. They, however, still require good initial estimates of camera poses (\eg, SAMURAI~\cite{boss2022samurai} uses manually annotated coarse camera poses as inputs) or an extremely large number ($>$50) of input images as the joint optimization of all unknown parameters easily falls into local minima.

\textbf{Pose estimation from specular reflection} has also been studied~\cite{lagger08reflections, schnieders13reflectionsonsphere, chang09specularcues, netzO13recognitionusingspecularhighlights,liu11posefromreflection,han21mirrorsurface}. Lagger \etal~\cite{lagger08reflections}, for instance, 
refine camera pose estimates using images and the 3D CAD model of a specular object. They recover view-dependent environment maps from the inputs and optimize the camera pose by minimizing discrepancies between them.
These methods, however, assume known geometry~\cite{lagger08reflections, schnieders13reflectionsonsphere, chang09specularcues, netzO13recognitionusingspecularhighlights}, known illumination~\cite{chang09specularcues,liu11posefromreflection,han21mirrorsurface}, or a simple lighting model~\cite{schnieders13reflectionsonsphere, netzO13recognitionusingspecularhighlights} which cannot be assumed in general scenes. 

In contrast, reflection correspondences enable us to recover the camera pose and object geometry of arbitrary objects under complex natural illumination without requiring static overlapping background to be present in the image.

\begin{figure*}[t]
    \centering
    \subfloat[][Pixel Correspondences]{
        \includegraphics[keepaspectratio, width=0.307\linewidth]{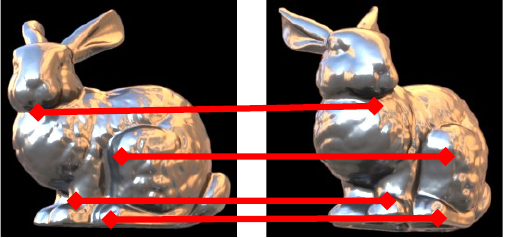}
        \label{fig:pixel-corrs}
    }
    \subfloat[][3D Correspondences]{
        \includegraphics[keepaspectratio, width=0.307\linewidth]{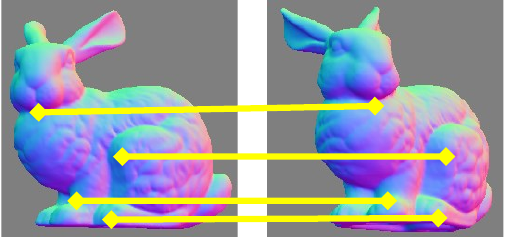}
        \label{fig:3D-corrs}
    }
    \subfloat[][Reflection Correspondences]{
        \includegraphics[keepaspectratio, width=0.307\linewidth]{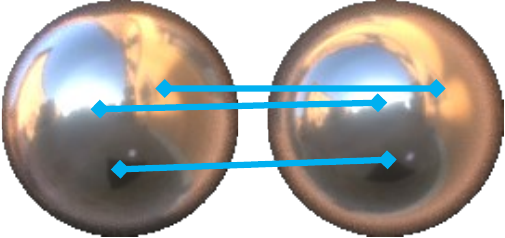}
        \label{fig:reflection-corrs}
    }
    \caption{We show how to leverage three types of correspondences. \emph{Pixel correspondences} (a) are the pixels that correspond to the same surface point. We can also leverage similar correspondences in the normal maps which we refer to as \emph{3D correspondences} (b). In addition to these correspondences, we leverage novel correspondences about the surrounding environment which we can observe through surface reflection. We recover camera-view reflectance maps, maps that associate surface normal orientations with the surrounding environment, and detect this type of correspondences from them (c). We refer to this novel kind of correspondences as \textbf{reflection correspondences}.}
    \label{fig:three_types}
\end{figure*}

\section{Method}

In this section, we first detail the three types of correspondences we consider and the equations that can be derived from them. \Cref{fig:three_types} provides a visualization of these three types of correspondences. To avoid confusion, we give distinct names to them.
\begin{itemize}
    \item ``pixel correspondences'' for the classical correspondences which link corresponding locations in the two images on the object surface: the two matched points are the reprojections of the same physical 3D point in the two images.
    \item ``3D correspondences'' for the correspondences between corresponding normals in the two images on the object surface. Note that for each pixel correspondence, we also have a 3D correspondence if we know the normals at the matched image locations. The difference is that for the pixel correspondence, we exploit the pixel coordinates themselves, while in the case of the 3D correspondence, we exploit the normals.
    \item ``reflection correspondences'' which match image locations where \emph{light rays mirror-reflected by the object surface come from the same direction in the two images}. As shown in \cref{fig:three_types}(c), we detect these correspondences from camera-view reflectance maps, \ie, view-dependent maps about the surrounding environment.
\end{itemize}

In this paper, we assume orthographic projection as objects are usually distant from the cameras and we can regard camera rays that point at the object as almost constant. Note that, as we see in the next subsection, camera pose estimation is challenging especially under orthographic projection.

\begin{figure*}[t]
    \centering
    \subfloat[][]{
        \includegraphics[keepaspectratio, height=0.23\linewidth]{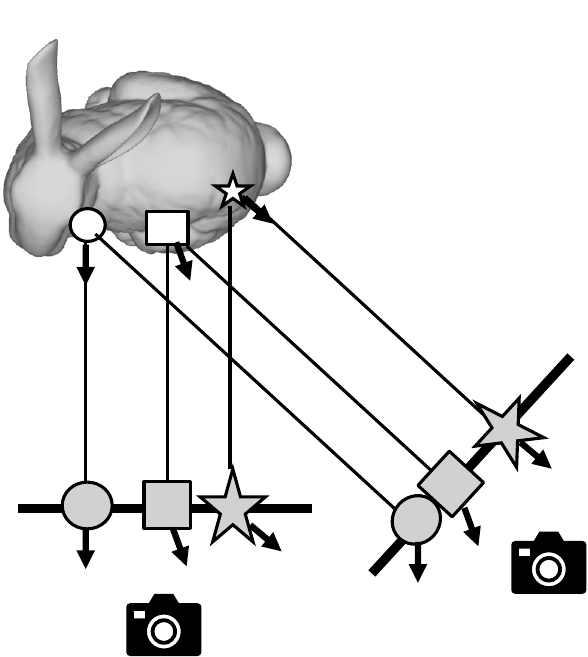}
        \label{fig:pose_ambiguity_a}
    }
    \subfloat[][]{
        \includegraphics[keepaspectratio, height=0.23\linewidth]{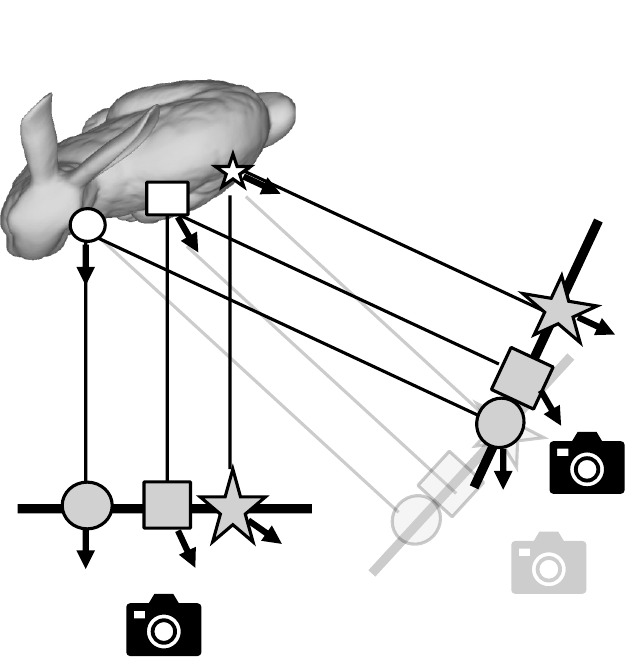}
        \label{fig:pose_ambiguity_b}
    }
    \subfloat[][]{
        \includegraphics[keepaspectratio, height=0.23\linewidth]{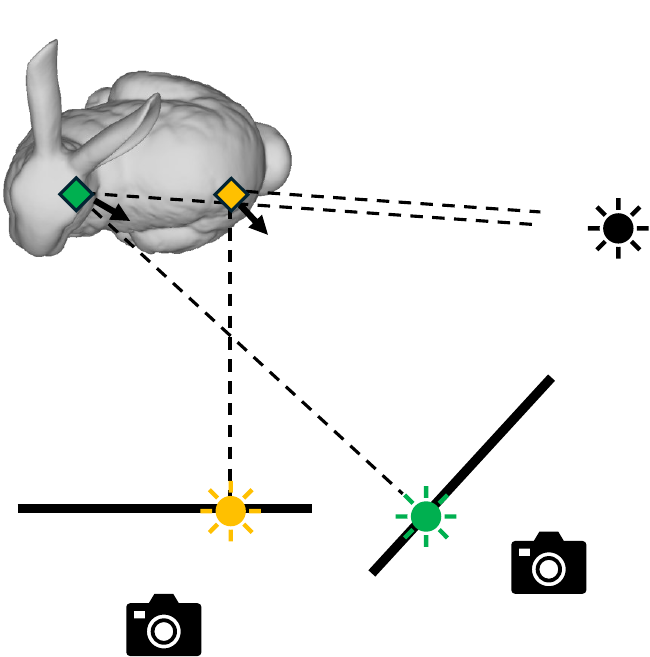}
        \label{fig:pose_ref_a}
    }
    \subfloat[][]{
        \includegraphics[keepaspectratio, height=0.23\linewidth]{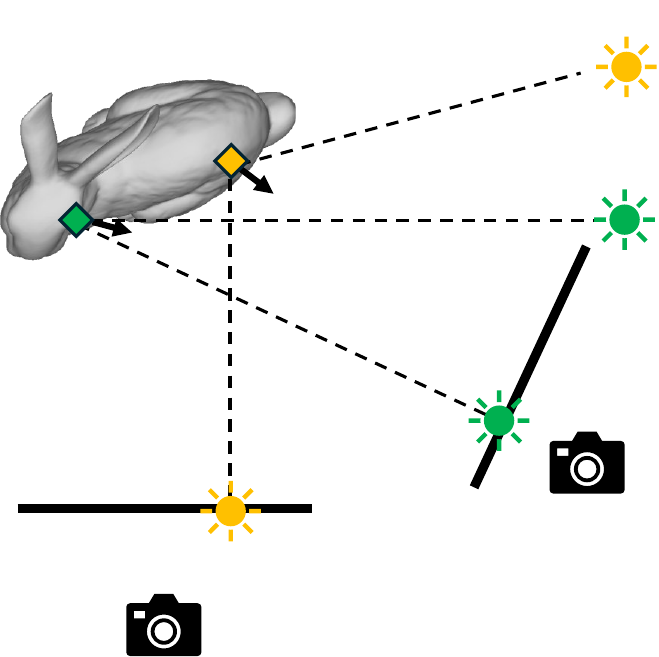}
        \label{fig:pose_ref_b}
    }
    \caption{
        As depicted in (\subref{fig:pose_ambiguity_a}) and (\subref{fig:pose_ambiguity_b}), due to the generalized bas-relief ambiguity in single-view surface normal recovery~\cite{belhumeur99basrelief} and the fundamental difficulty in structure-from-motion~\cite{harris91orthographicsfm}, we cannot obtain a unique solution for the relative rotation from pixel and 3D correspondences when the cameras are orthographic (\eg, distant from the object). Reflection correspondences, \ie, correspondences regarding the incident directions for specular reflections, enable us to distinguish the correct relative rotation (\subref{fig:pose_ref_a}) from the other possible solutions like (\subref{fig:pose_ref_b}).
    }
    \label{fig:ambiguity}
\end{figure*}

\subsection{Pixel Correspondences}
Let us assume that we can establish $N_\IM$ pixel correspondences $(u_1^i, v_1^i)\leftrightarrow (u_2^i, v_2^i)$. Under orthographic projection, each pixel correspondence gives us one constraint on the camera rotation between the two images~\cite{harris91orthographicsfm}:
\begin{equation}
   \left[ (u_1^i \cos\phi + v_1^i \sin \phi) - (u_2^i \cos\theta + v_2^i \sin \theta)\right]^2 = 0\,,
    \label{eq:image_correspondences}
\end{equation}
where $\phi$ and $\theta$ are two of the three Euler angles $(\theta, \phi, \eta)$ that correspond to the relative rotation $\mathrm{R_{21}}$ from the second view to the first one. Without loss of generality, we assume that these correspondences are translated such that the average location (\eg, $\frac{1}{N_\IM}\sum_i u_1^i$) is zero. We use the (z-x-z) sequence for the Euler angles
\begin{equation}
    \mathrm{R_{21}} = \mathrm{R_z}[\phi]~\mathrm{R_x}[\eta]~\mathrm{R_z}[-\theta]\,,
\end{equation}
where $\mathrm{R_z}[\cdot]$ and $\mathrm{R_x}[\cdot]$ are rotations around z and x axes, respectively.
As depicted in \cref{fig:pose_ambiguity_a,fig:pose_ambiguity_b}, since \cref{eq:image_correspondences} is independent of one of the unknown Euler angles ($\eta$), the rotation matrix cannot be fully recovered from pixel correspondences of only two views. 

\subsection{3D Correspondences}

If we could estimate unambiguously the normal maps, $\bN_1$ and $\bN_2$, for the images, correspondences $(u_1^i, v_1^i)\leftrightarrow (u_2^i, v_2^i)$ would be constrained by 
\begin{equation}
    \bN_1[u_1^i,v_1^i] = \mathrm{R_{21}}\bN_2[u_2^i,v_2^i]\,,
    \label{eq:normal_map_correspondences_wo_gbr}
\end{equation}
as the normals at locations $(u_1^i, v_1^i)$ and $(u_2^i, v_2^i)$ are the same up to rotation $\mathrm{R}_{21}$. 
We then could estimate the relative rotation $\mathrm{R}_{21}$ by simply solving \cref{eq:normal_map_correspondences_wo_gbr}. 

Unfortunately, in practice, normal maps can be recovered only up to the generalized bas-relief~(GBR) ambiguity~\cite{belhumeur99basrelief}. Given a normal map $\mathbf{N}$ for an image $I$, any normal map $\bN'$ with 
\begin{equation}
    \bN'[u,v] = \mathrm{G}^{-\transp} \bN[u,v] \,,
    \label{eq:bas_relief_nm}
\end{equation}
could result in the same image $I$ under a different lighting, where $ \mathrm{G}$ is the generalized bas-relief~(GBR) transformation~\cite{belhumeur99basrelief}
\begin{equation}
    \mathrm{G} \equiv 
    \begin{pmatrix}
        1 & 0 & 0 \\
        0 & 1 & 0 \\
        \mu & \nu & \lambda
    \end{pmatrix}
    \,.
\end{equation}
The parameters $\mu$ and $\nu$ can take any values, and $\lambda$ can take any positive value. Note that, although discussions regarding the bas-relief ambiguity (\eg, one in Belhumeur \etal~\cite{belhumeur99basrelief}) usually assume directional lights, this holds true even for environmental illumination as we can view it as a set of directional lights.

The bas-relief ambiguity changes the relationship in \cref{eq:normal_map_correspondences_wo_gbr} to 
\begin{equation}
    \bN_1[u_1^i,v_1^i] \propto \mathrm{G_1}^{-\transp}\mathrm{R_{21}}\mathrm{G_2}^\transp\bN_2[u_2^i,v_2^i]\,,
    \label{eq:normal_map_correspondences}
\end{equation}
where $\mathrm{G_k}$ $(k=1,2)$ is an unknown GBR transformation for each view that corresponds to the estimation errors~\cite{belhumeur99basrelief}. If we have a sufficient number of correspondences, we can obtain a unique solution for the combined transformation
\begin{equation}
    \mathrm{G}_{21} \equiv \mathrm{G_1}^{-\transp}\mathrm{R_{21}}\mathrm{G_2}^\transp\,.
\end{equation}
There is, however, an unresolvable ambiguity in its decomposition into the three unknown matrices. Most important, this ambiguity corresponds to one regarding the pixel correspondences. In other words, we cannot recover the relative camera pose even when combining pixel and 3D correspondences: for any $\eta$, there are corresponding GBR transformations $\mathrm{G_1}$ and $\mathrm{G_2}$ that are consistent with \cref{eq:normal_map_correspondences}. We provide the proof in the supplementary material.

\subsection{Reflection Correspondences}

Let us assume we have correspondences between image locations $(u_1^i, v_1^i)\leftrightarrow (u_2^i, v_2^i)$ where \emph{light rays come from the same direction in the two images}. In other words, correspondences of the surroundings reflected by the object surface.
If we, for now, ignore the GBR ambiguity, these ``reflection correspondences'' give each a constraint of the form
\begin{equation}
    \refl(\bN_1[u_1^i, v_1^i]) \propto \mathrm{R_{21}} \refl(\bN_2[u_2^i, v_2^i]) \,, 
    \label{eq:reflectance_map_correspondences_wo_gbr}
\end{equation}
where function $\refl(\bn)$ returns the reflection of the line of sight direction $\omega_o$ on the surface with normal $\bN$:
\begin{equation}
    \refl(\bn) = -\omega_o + 2 \frac{\omega_o \cdot \bn}{\bn\cdot\bn}\bn\,.
\end{equation} 
Again, at this stage, we can in fact predict the normal maps only up to the GBR ambiguity, and we need to introduce the two GBR transformations into \cref{eq:reflectance_map_correspondences_wo_gbr}:
\begin{equation}
    \refl(\mathrm{G_1}^\transp \bN_1[u_1^i, v_1^i]) \propto \mathrm{R_{21}} \refl(\mathrm{G_2}^\transp \bN_2[u_2^i, v_2^i]) \,.
    \label{eq:reflectance_map_correspondences}
\end{equation}
As illustrated in \cref{fig:pose_ref_a,fig:pose_ref_b}, each reflection correspondence thus gives us a new type of equation to estimate $\mathrm{R_{21}}$ but also $\mathrm{G_1}$ and $\mathrm{G_2}$, from which we can also get the object shape. 

Detecting these reflection correspondences directly from images is, however, extremely challenging as surface reflection depends not only on the surrounding illumination but also on the surface geometry. Let us now assume we have a surface normal map for each view. Then we can avoid this problem by recovering camera-view reflectance maps~\cite{horn1979reflectancemap}. The reflectance maps are view-dependent mappings from a surface normal to the surface radiance which are determined by the surface reflectance and the surrounding illumination environment
\begin{equation}
    E_k(\mathbf{n}) = \int L_i(\mathrm{R}_k^\mathrm{T}\mathbf{\omega_i}) \psi(\mathbf{\omega_i}, \mathbf{\omega_o}, \mathbf{n}) \max(\mathbf{\omega_i} \cdot \mathbf{n}, 0)\mathrm{d}\mathbf{\omega_i}\,,
\end{equation}
where $\mathbf{\omega_i}$, $\mathbf{\omega_o}$, and $\mathbf{n}$ are incident, viewing, and surface normal orientations in the local camera coordinate system, respectively. $\mathrm{R}_k$ is the camera pose (a rotation matrix) of $k$-th view, $L_i(\mathrm{v})$ is a mapping from an incident direction to a radiance of the corresponding incident light, and $\psi(\mathbf{\omega_i}, \mathbf{\omega_o}, \mathbf{n})$ is the bidirectional reflectance distribution function (BRDF). 

From the surface normal maps and the input images, \ie, pairs of surface normals and surface radiances, we can recover these reflectance maps~\cite{rematas16drm,yamashita2023deepsharm} and, as illustrated in \cref{fig:three_types}(c), detect these reflection correspondences from the reflectance maps regardless of the object shape. Note that, as the reflectance maps are recovered using the surface normal maps, they also suffer from the bas-relief ambiguity. The relationship between the ground truth $E(\mathbf{n})$ and another possible solution $E'(\mathbf{n})$ is 
\begin{equation}
    E'(\mathbf{n}) = E(\mathrm{G}^\mathrm{T} \mathbf{n})\,.
    \label{eq:bas_relief_rm}
\end{equation}
Thus we still need to use \cref{eq:reflectance_map_correspondences} for these reflection correspondences.

\subsection{Relative Rotation from Correspondences}
\label{sec:objective}

From \cref{eq:image_correspondences}, \eqref{eq:normal_map_correspondences}, and \eqref{eq:reflectance_map_correspondences}, we can derive an objective function
\begin{equation}
    f = f_\IM + f_\NM^{(12)} + f_\NM^{(21)} + f_\RM^{(12)} + f_\RM^{(21)} \,.
    \label{eq:objective}
\end{equation}
that enforces the equation in the least-squares sense. $f_\IM$ enforces \cref{eq:image_correspondences}:
\begin{equation}
    f_\IM = \frac{1}{{N_\IM}}\sum_i^{N_\IM} \left( t_\phi^i - t_\theta^i \right)^2\,,
    \label{eq:obj_image_correspondences}
\end{equation}
with
\begin{equation}
    t_\phi^i = u_1^i \cos\phi + v_1^i \sin \phi\,,
\end{equation}
\begin{equation}
    t_\theta^i = u_2^i \cos\theta + v_2^{(i)} \sin \theta\,,
\end{equation}
and $N_\IM$ is the number of image correspondences.

Note that for \cref{eq:normal_map_correspondences} and \cref{eq:reflectance_map_correspondences}, we can switch the role of the two images. We therefore introduce two terms for each of these equations. For \cref{eq:normal_map_correspondences}, we introduce
\begin{equation}
    f_\NM^{(jk)} = \frac{1}{{N_\NM}}\sum_i^{N_\NM} \left\|\bN_j[u_j^i, v_j^i]  - \Norm(\mathrm{G}_{kj} \bN_k[u_k^i, v_k^i])\right\|^2\,,
    \label{eq:objective_nm}
\end{equation}
where $\Norm$ is the vector normalization operator and $N_\NM$ the number of 3D correspondences. 

For \cref{eq:reflectance_map_correspondences}, we introduce
\begin{equation}
    f_\RM^{(jk)} = \frac{1}{{N_\RM}}\sum_i^{N_\RM} \left\| \bN_j[u_j^i, v_j^i] -  
    \Omega^{(jk)}( \bN_k[u_k^i, v_k^i] ) \right\|^2\,,
    \label{eq:objective_rm}
\end{equation}
where $\Omega^{(jk)}(\bn)$ transforms a surface normal $\bn$ in the $j$-th view reflectance map to the surface normal in the $k$-th view
\begin{equation}
    \Omega^{(jk)}(\bn) = \Norm( \mathrm{G}_k^{-T}\refl^{-1}(\mathrm{R}_{jk} \; \refl(\mathrm{G}_j^T \bn )))\,.
\end{equation}

Given 3 sets of $N_\IM$ pixel correspondences, $N_\NM$ 3D correspondences, and $N_\RM$ reflection correspondences, we can optimize $f$ in \cref{eq:objective} for the three Euler angles $\theta$, $\phi$, and $\eta$, and the two sets of parameters for the GBR transformations for both images $\mu_1$, $\nu_1$, $\lambda_1$, $\mu_2$, $\nu_2$, $\lambda_2$, under the constraints that $\lambda_1$ and $\lambda_2$ are positive. 

Naive optimization with all the correspondences can easily be affected by erroneous correspondences which are unavoidable. We also empirically find that it is slow and susceptible to local minima especially due to the nonlinear operation $\refl(\cdot)$. We instead derive a RANSAC-based, two-step algorithm that first estimates the combined transformation $\mathrm{G_{21}}$ using pixel and 3D correspondences and then decomposes it using reflection correspondences.

In the first step, given the pixel and the 3D correspondences, our algorithm recovers the combined transformation $\mathrm{G}_{21} \equiv \mathrm{G_1}^{-\transp}\mathrm{R_{21}}\mathrm{G_2}^\transp$ based on RANSAC~\cite{fischler81ransac}. For each of the $L_1$ iterations, we build a set consisting of randomly sampled $M_1$ pixels and $M_1$ 3D correspondences. In practice, we set $M_1$ to 4. We empirically find this to be the smallest necessary number of pixel and 3D correspondences. Please see the supplementary material for details. We hope to derive a theoretical justification of this number in our future work. For each set, we obtain an estimate of $\mathrm{G}_{21}$ by minimizing an objective function $f_1$ using an off-the-shelf solver for nonlinear optimization~\cite{2020SciPy-NMeth,branch99trf}. $f_1$ is similar to the sum of $f_\IM$, $f_\NM^{(12)}$, and $f_\NM^{(21)}$ in \cref{eq:objective}, though we compute $f$ using only the sampled correspondences. We then select a good estimate $\hat{\mathrm{G}}_{21}$ from them according to the number of pixels and 3D correspondences that are consistent with the estimate.

In the second step, given the estimate $\hat{\mathrm{G}}_{21}$ and the reflection correspondences, we obtain a combination of $\mathrm{G}_1$, $\mathrm{G}_2$, and $\mathrm{R}_{21}$ that is consistent with $\hat{\mathrm{G}}_{21}$ and most of the reflection correspondences. The key idea here is that, if we determine $\eta$, one of the three Euler angles of $\mathrm{R}_{21}$ that remains ambiguous, we can decompose $\hat{\mathrm{G}}_{21}$ into $\mathrm{G}_1$, $\mathrm{G}_2$, and $\mathrm{R}_{21}$ uniquely and analytically. Based on this, for each possible $\eta$, we compute the corresponding $\mathrm{G}_1$, $\mathrm{G}_2$, and $\mathrm{R}_{21}$, and select a solution that maximizes the number of reflection correspondences that are consistent with the decomposed transformations. 
The supplementary material provides more details and pseudo code.

\subsubsection{Translation Estimation}

Once $\mathrm{R_{21}}$ is recovered, by leveraging inlier pixel correspondences, we can also solve for the relative translation between the two views except for offsets regarding the two viewing directions~\cite{harris91orthographicsfm}. That is, we estimate the translation vector $\mathbf{t_{21}}=(t_x, t_y, t_z)$ using the constraint
\begin{equation}
    r_{23} t_x - R_{13} t_y = -\sum_i \left(r_{23}\tilde{u}^i - r_{13}\tilde{v}^i\right)\,,
\end{equation}
where $r_{jk}$ is $(j,k)$ element of $\mathrm{R_{21}}$ and
\begin{equation}
    \tilde{u}^i = r_{11} u_2^i + r_{12} v_2^i - u_1^i\,,
\end{equation}
\begin{equation}
    \tilde{v}^i = r_{21} u_2^i + r_{22} v_2^i - v_1^i\,.
\end{equation}

\begin{figure*}[t]
  \centering
  \includegraphics[keepaspectratio, width=\linewidth]{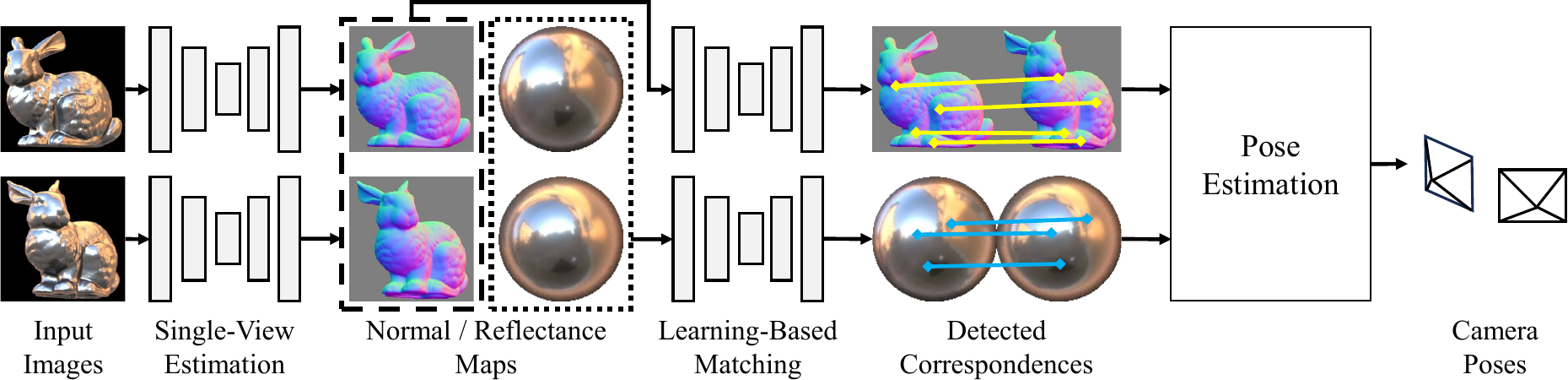}
  \caption{Given two-view images of a textureless, non-Lambertian object, we first recover the surface normals and a reflectance map for each view using a single-view geometry reconstruction method~\cite{yamashita2023deepsharm}. We establish 3D and reflection correspondences with a novel deep feature extraction network to compute the relative camera pose from them.
  }
  \label{fig:method-overview}
\end{figure*}

\subsection{Camera Pose Estimation from Two Images}

Based on the algorithm in \cref{sec:objective}, we can derive a method, as depicted in \Cref{fig:method-overview}, that takes in only two-view images of a textureless, non-Lambertian object and estimates the relative camera pose (the relative rotation and the relative translation) between the two views. We achieve this by recovering a surface normal and a reflectance map for each view, detecting the correspondences from the recovered maps, and applying the proposed algorithm to the detected correspondences. 

To recover the normal and reflectance maps for each view separately, we use DeepShaRM~\cite{yamashita2023deepsharm} which is a radiometry-based geometry estimation method. DeepShaRM can jointly recover geometry, reflectance maps, and normal maps from an image by alternating between (1) learning-based reflectance map estimation from images and a geometry estimate, (2) learning-based surface normal estimation from images and the estimated reflectance maps, and (3) geometry optimization with the estimated surface normals. This gives us the normal and reflectance maps but only up to the GBR transformations. Please see the supplementary material for more implementation details.

To obtain 3D correspondences in the normal maps, we train a convolutional neural network to extract pixel-wise, view-invariant features that we use to match surface points between normal maps. 
We also train another deep neural network for the detection of reflection correspondences in the same way.
We train the network by contrastive learning. We use pairs of synthetic normal maps of two views as training data. We feed both of them to the feature extraction network and obtain corresponding feature maps $\bF_1$ and $\bF_2$. Given ground truth correspondences $(u_1^i, v_1^i) \leftrightarrow (u_2^i, v_2^i)$, we impose the InfoNCE loss~\cite{oord18infonce} 
\begin{equation}
    L = \sum_i\sum_j-\delta_{ij}\log\left(\frac{ \exp(c_{ij})}{\sum_{i'}\sum_{j'} \exp(c_{i'j'})}\right)\,,
    \label{eq:infonce}
\end{equation}
where $c_{ij}$ is the cosine similarity of feature vectors $\bF_1[u_1^i, v_1^i]$ and $\bF_2[u_2^j, v_2^j]$
\begin{equation}
  c_{ij} = \Norm(\bF_1[u_1^i, v_1^i]) \cdot \Norm(\bF_2[u_2^j, v_2^j])\,,
\end{equation}
and $\delta_{ij}$ is the Kronecker delta.
By this, we ensure that features of pixels that correspond to the same surface point become similar.

The training with synthetic normal maps is, however, insufficient in practice as the surface normal maps estimated by the single-view estimation are distorted by GBR transform. We overcome this with data augmentation. For each view in the training data, we randomly sample parameters of the GBR transformation and transform the input normal map according to \cref{eq:bas_relief_nm}. We use the transformed normal maps instead of the original ones so that the network can learn to extract features robust to the estimation errors caused by the bas-relief ambiguity. 

At inference time, using the extracted feature maps, we detect correspondences by brute-force matching and filter them with the ratio test by Lowe~\cite{lowe04imagefeature}. For any location on the object in the first image (or reflectance map), we look for the best and second best locations in the second image (or reflectance map) in terms of the cosine similarity of the features. If the ratio of the similarity of the second best location to one of the best location is lower than a threshold, we use the best location as the matched location. This gives us all of pixel, 3D, and reflection correspondences.

Once we have the three types of correspondences, we can recover the relative rotation and the relative translation using the two-step estimation algorithm (\cref{sec:objective}).

\subsection{Joint Shape and Camera Pose Recovery}
\label{sec:joint_est}

Once we obtain the relative camera pose, we can further improve its accuracy by consolidating the multi-view surface normal estimates using the camera pose estimates and exploiting the improved surface normals and corresponding reflectance maps to update the camera pose estimate. As DeepShaRM~\cite{yamashita2023deepsharm} is originally designed for posed multi-view images, we achieve this by alternating between multi-view surface normal and reflectance map estimation by DeepShaRM and camera pose estimation by our method. 
As a byproduct of DeepShaRM, we can also obtain an accurate object shape.

\section{Experimental Results}
\label{sec:results}

\begin{table}[t]
  \centering
  \setlength{\tabcolsep}{2.7pt}
  \def\HEAD#1#2{\begin{tabular}{c}#1\\#2\end{tabular}}
  \scriptsize

  \caption{
    (\subref{tab:camera-pose-results}) Mean camera pose (relative rotation) estimation accuracy on images in the nLMVS-Synth dataset~\cite{yamashita2023nlmvs}.
    (\subref{tab:camera-pose-results-real}) Camera pose estimation accuracy on real-world images. (*) COLMAP~\cite{schoenberger2016colmap_sfm} uses 11 view uncropped images as inputs. 
  }
  
    \centering
    \subfloat[][Synthetic Result]{
      \begin{tabular}{l|c}
         & Pose Error \\ \hline
        w/o Data Augm. & 7.4 deg   \\
        w/o Joint & 6.2 deg \\
        w/o RM & 6.6 deg \\
        \rowcolor[rgb]{0.93,1.0,0.87} Ours & 4.5 deg
      \end{tabular}
      \label{tab:camera-pose-results}
    }
    ~
    \subfloat[][Real Data Result]{
      \begin{tabular}{l|ccccc}
         & Planck & Horse & Bunny & Cat \\ \hline
         COLMAP~\cite{schoenberger2016colmap_sfm} (*) & 15.7 deg & N/A & 24.5 deg & N/A \\
         LightGlue~\cite{lindenberger23lightglue} & 5.0 deg & 18.9 deg & 93.6 deg & 35.4 deg \\
         SAMURAI~\cite{boss2022samurai} & 5.6 deg & 24.3 deg & 30.5 deg & 68.7 deg \\
        \rowcolor[rgb]{0.93,1.0,0.87} Ours & 1.9 deg & 12.5 deg & 7.2 deg & 8.6 deg \\
      \end{tabular}
      \label{tab:camera-pose-results-real}
    }
  \setlength{\tabcolsep}{2.7pt}
  
\end{table}

We focus our experiments on answering the following key questions. 
How well do the reflection correspondences contribute to the camera pose estimation accuracy?
Can the novel feature extraction network establish correspondences robust to GBR distortion? 
How does our method generalize to real-world images? 
How does our method compare to structure-from-motion methods and neural image synthesis methods with camera pose optimization?
To answer these questions, we validate the following points
\begin{itemize}
    \item The proposed components (especially the reflection correspondences) are essential for accurate reconstruction from a sparse set (2) of images.
    \item Our method can recover accurate camera poses even from two real images.
\end{itemize}

\textbf{Training Data} We trained the deep networks in our method (including DeepShaRM~\cite{yamashita2023deepsharm}) on the training set of the nLMVS-Synth dataset~\cite{yamashita2023nlmvs}. The training set consists of 26850 images of 2685 synthetic objects. The training shapes are composed of primitive shapes (ellipsoids, cubes, cylinders) augmented with random height fields~\cite{xu2018relighting}. 94 materials and 2685 environmental maps from existing databases~\cite{matusik2003brdf,gardner2017indoor,hdrihaven} are used for rendering.

\begin{figure*}[t]
    \centering
    \subfloat[][]{
        \includegraphics[keepaspectratio, height=0.3\linewidth]{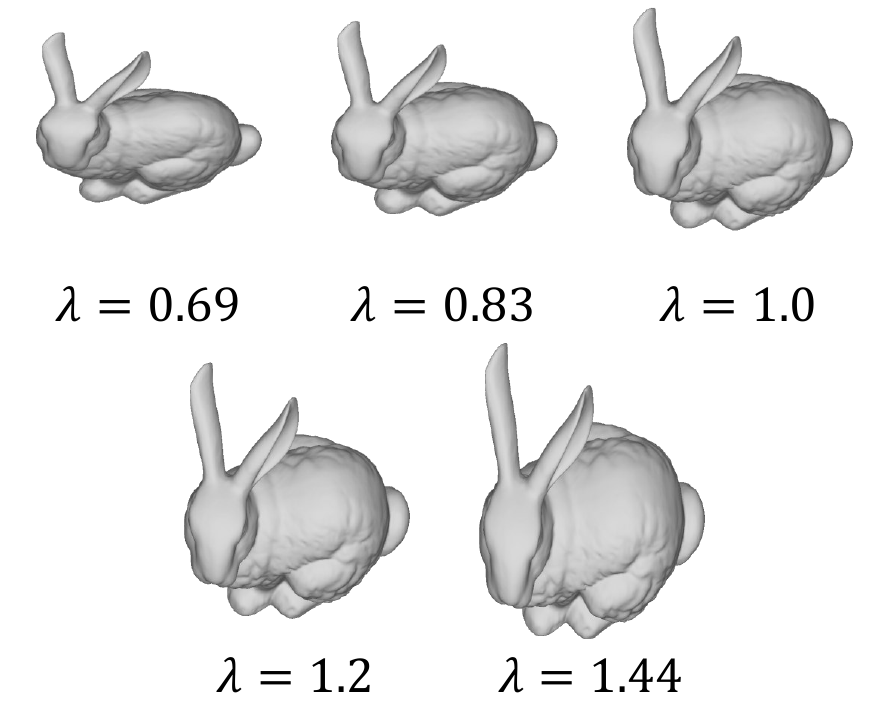}
        \label{fig:distorted_bunnies}
    }
    \subfloat[][]{
        \includegraphics[keepaspectratio, height=0.3\linewidth]{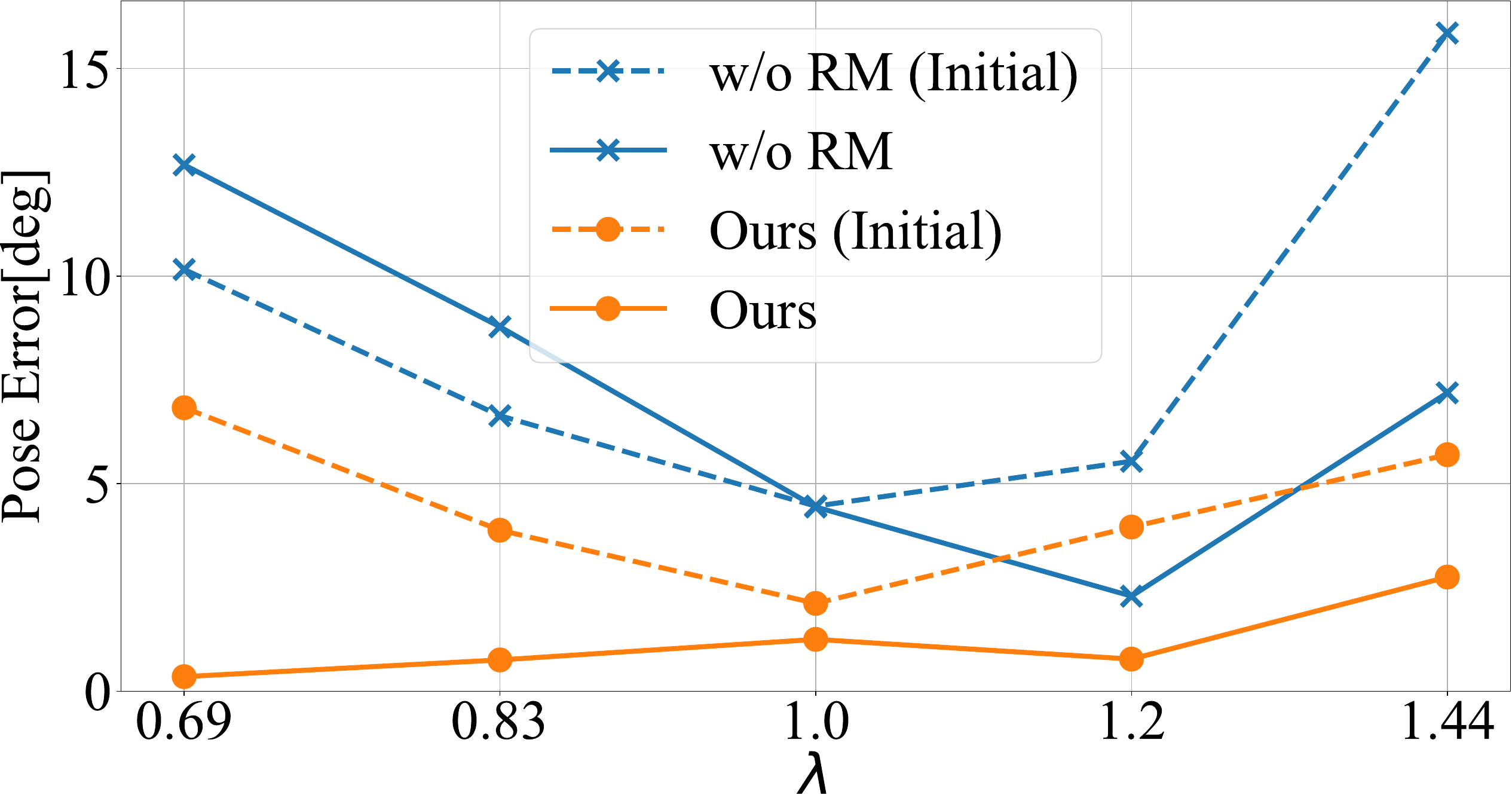}
        \label{fig:result_lambda_vs_accuracy}
    }
    \caption{
        Evaluation on synthetic shapes with different levels of flatness. (\subref{fig:distorted_bunnies}) We create such shapes by applying GBR transformations with different $\lambda$. (\subref{fig:distorted_bunnies}) Accuracy for each $\lambda$. The results shows the effectiveness of using reflection correspondences. Please see the text for details.
    }
    \label{fig:result_ablation_flatness}
\end{figure*}

\subsection{Evaluation on Synthetic Data}

We evaluate the accuracy of the joint camera pose and object shape estimation framework on synthetic images rendered with unseen shapes, BRDFs, and illumination maps. We used synthetic images from the test set of the nLMVS-Synth dataset~\cite{yamashita2023nlmvs} and some rendered by ourselves for this evaluation. In total, the test set consists of 27 combinations of 3 shapes, 3 BRDFs, and 3 illumination environments.

We evaluate the accuracy of our method by computing the geodesic distance between the estimated relative rotation and the ground truth. Note that evaluation of the estimated translation under orthographic projection is difficult due to unsolvable offsets regarding the two viewing directions.
We compare our camera pose estimation accuracy with those by its own ablated variants, ``w/o Data Augm.'', ``w/o Joint,'' and ``w/o RM.''. ``w/o Data Augm.'' uses the feature extraction networks trained without the data augmentation with GBR transforms. ``w/o Joint'' is our method without the joint iterative estimation which recovers camera poses from only the initial estimates of normal and reflectance maps. ``w/o RM'' ignores the bas-relief ambiguity and recovers the relative rotation using only pixel and 3D correspondences based on \cref{eq:image_correspondences,eq:normal_map_correspondences_wo_gbr}.

\Cref{tab:camera-pose-results} shows the average camera pose estimation errors. The results show that the joint estimation, the data augmentation method, and the use of reflection correspondences are essential for accurate camera pose estimation.

\subsection{Robustness to Various Levels of Flatness}

To further clarify the effectiveness of the reflection correspondences, we also test our method on synthetic shapes with different levels of flatness. As shown in \Cref{fig:distorted_bunnies}, we created such test shapes by applying GBR transformations with different parameters to the Stanford Bunny~\cite{stanford3dscan}. We set $\mu$ and $\nu$ in the GBR transformations to be zero and $\lambda$ to be one of 0.69, 0.83, 1.0, 1.2, and 1.44. Using the distorted shapes along with BRDF data and illumination maps used in the synthetic evaluation above, we rendered synthetic images for this evaluation. We tested our method and the baseline method that does not exploit reflection correspondences (``w/o RM'') on them.

\Cref{fig:result_lambda_vs_accuracy} shows the mean camera pose estimation error for each $\lambda$, \ie, each level of flatness. The baseline method that exploits only conventional correspondences works well only for a shape with a ``normal'' level of flatness (\ie, close to $\lambda=1$). This is because the method heavily relies on a realistic prior learned by the deep geometry estimation method~\cite{yamashita2023deepsharm}. In contrast, our method is robust to various levels of flatness which is critical for practical use in the real-world.

\begin{figure*}[t]
  \centering
  \includegraphics[keepaspectratio, width=\linewidth]{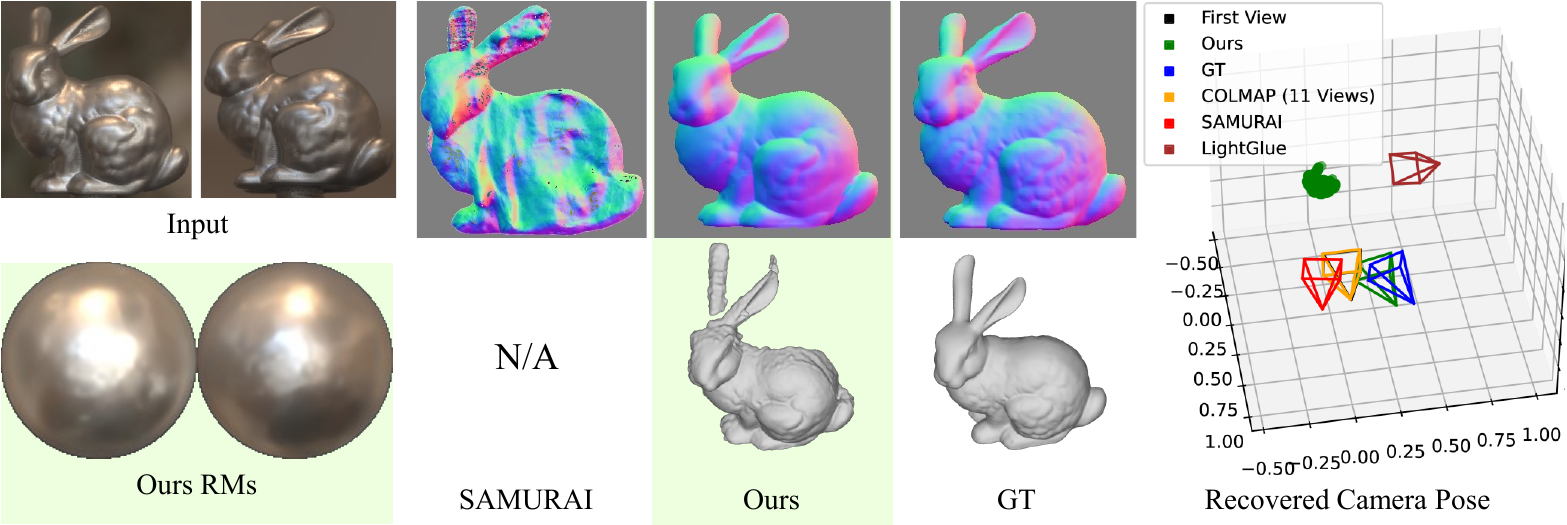}
  \caption{Reflectance maps (RMs), normal maps, surface geometry, and relative camera poses recovered from two-view real-world images~\cite{yamashita2023nlmvs}.
  }
  \label{fig:result_real}
\end{figure*}

\begin{figure*}[t]
  \centering
  \includegraphics[keepaspectratio, width=0.8\linewidth]{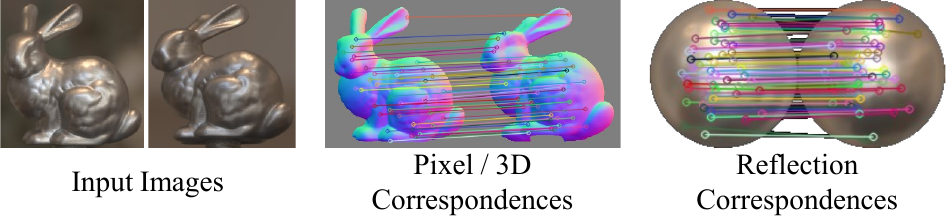}
  \caption{All types of correspondences detected from the real image pair for \cref{fig:result_real}.
  }
  \label{fig:result_real_corr}
\end{figure*}

\subsection{Evaluation on Real Data}

Quantitative evaluation of our method on in-the-wild images is difficult as they are by definition difficult to register with traditional methods. Nevertheless, we were able to evaluate our method on the real images from the nLMVS-Real dataset~\cite{yamashita2023nlmvs} and some captured and calibrated by ourselves.

We applied our method trained on the synthetic data to images of four real objects without any fine tuning. We compared our results with those of applying COLMAP~\cite{schoenberger2016colmap_sfm}, LightGlue~\cite{lindenberger23lightglue}, and SAMURAI~\cite{boss2022samurai}. We found that COLMAP completely fails on our inputs, \ie, two-view cropped images that cover only the target object. For images from the nLMVS-Real, we used 11 view uncropped images that capture not only the target object but also textured object around the target (\ie, ChArUco boards) as inputs to COLMAP. Note that the uncropped images are provided by the authors of the dataset~\cite{yamashita2023nlmvs} as RAW data.

\Cref{tab:camera-pose-results-real} and \cref{fig:result_real} show quantitative and qualitative results. Note that COLMAP completely failed on the ``Horse'' object. As SAMURAI failed to extract a 3D mesh model from their volumetric geometry representation, we only show a normal map for SAMURAI. In contrast to these existing methods which fail on these challenging inputs, our method successfully recovers plausible camera poses, surface geometry, and reflectance maps.

\subsubsection{Scenes with textured and textureless objects}
We can also leverage our method to estimate camera poses from sparse images capturing scenes consisting of both textured and textureless, non-Lambertian objects often found in the real world. In such a setting, the pixel correspondences can come from the textured object and the reflection correspondences from the textureless object. That is, the different types of correspondences can come from different objects in the scene to recover the unique camera poses capturing all the objects 
\begin{wrapfigure}[19]{r}{0.45\textwidth}
    \centering
    \includegraphics[width=0.95\linewidth]{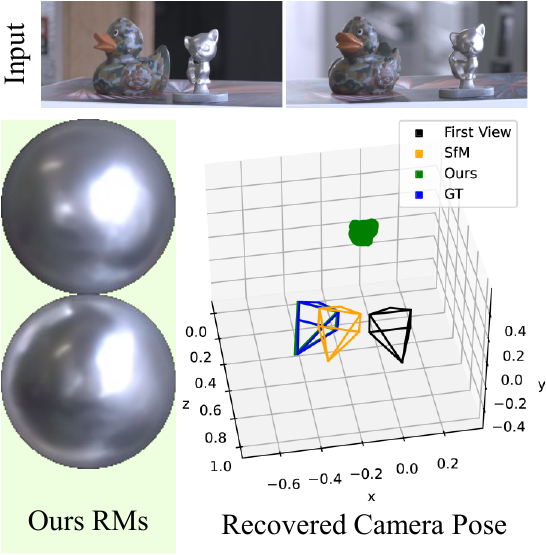}
    \caption{
    Results on images capturing scenes consisting of both textured and textureless, non-Lambertian objects. Our result aligns almost perfectly with the ground truth.
    }
    \label{fig:result_real_scene}
\end{wrapfigure}
in each shot.

\Cref{fig:result_real_scene} shows qualitative result of our method and that of a baseline structure-from-motion (SfM) method on an example scene. Note that, although the cameras are almost orthographic, we can still use a perspective model to obtain a unique estimate of the relative rotation from only the pixel correspondences. The results show that, while the SfM method (orange) struggles to recover the angle between the two views (\ie, $\eta$), our camera pose estimation result (green) is almost identical to the ground truth (blue). This demonstrates the effectiveness of our method even in scenes consisting of multiple different types of objects.  Please see the supplementary material for more results.

\section{Conclusion}

We introduced a new type of correspondences, reflection correspondences, which enables accurate and robust camera pose and joint shape estimation from the appearance alone of textureless, reflective objects. We believe this new type of correspondences not only expunges restricting requirements on image capture for camera pose estimation but also opens new use of object appearance in various applications. We hope to explore these in future work and also catalyze research for this with our release of code and data.

\subsubsection*{Acknowledgements}
This work was in part supported by
JSPS KAKENHI 
20H05951 and 
21H04893; 
JST JPMJCR20G7, 
JPMJAP2305, 
and JPMJSP2110; 
RIKEN GRP;
ANR project TOSAI ANR-20-IADJ-0009; 
and the ERC grant "explorer" (No.~101097259).

%
%
\bibliographystyle{splncs04}
\bibliography{kyamashita_pose_camera_ready}

\appendix
\numberwithin{equation}{section}
\renewcommand{\thetable}{\Alph{section}.\arabic{table}}
\renewcommand{\thefigure}{\Alph{section}.\arabic{figure}}

\section{Evaluation on Synthetic Data}

\begin{figure}[t]
  \centering
  \includegraphics[keepaspectratio, width=\linewidth]{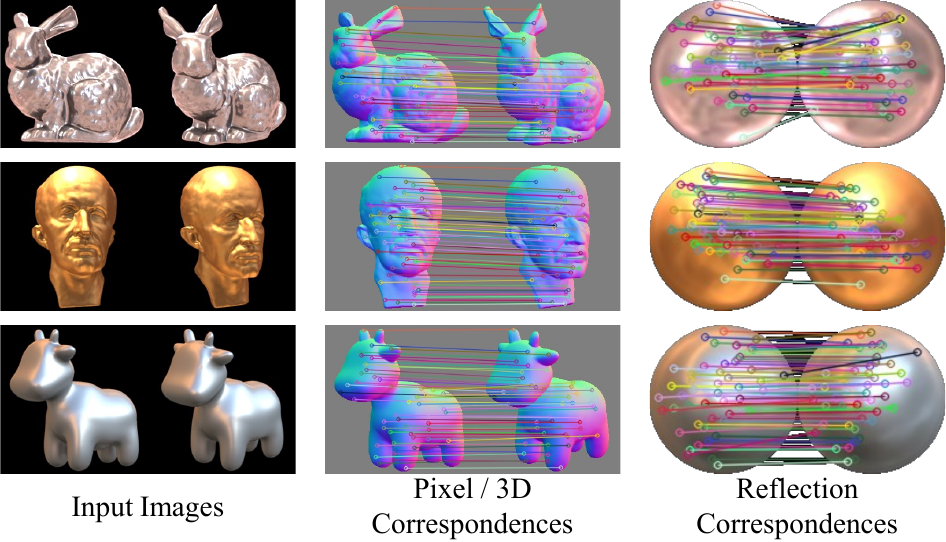}
  \caption{Examples of correspondences detected from the synthetic test images. Visual inspection shows that they match semantically correct surface and reflectance map points.}
  \label{fig:result_corrs_synth}
\end{figure}

\Cref{fig:result_corrs_synth} shows inlier correspondences detected by our method. Most of them join points on the surface and reflectance maps that visually appear to be in correspondence. These results demonstrate the effectiveness of the feature extractor and the robust estimation method. Their accuracy is ultimately reflected in the accuracy of the resulting camera pose estimates.

\begin{figure*}[t]
    \centering
    \subfloat[][Accuracy for each illumination]{
        \includegraphics[keepaspectratio, width=0.45\linewidth]{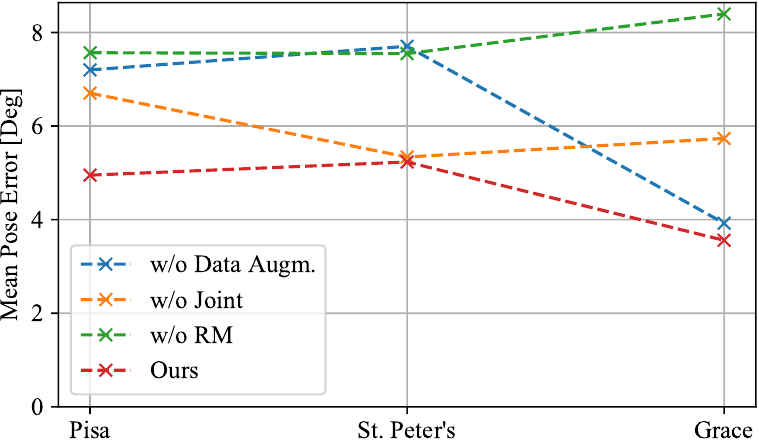}
    }
    \subfloat[][Accuracy for each material]{
        \includegraphics[keepaspectratio, width=0.45\linewidth]{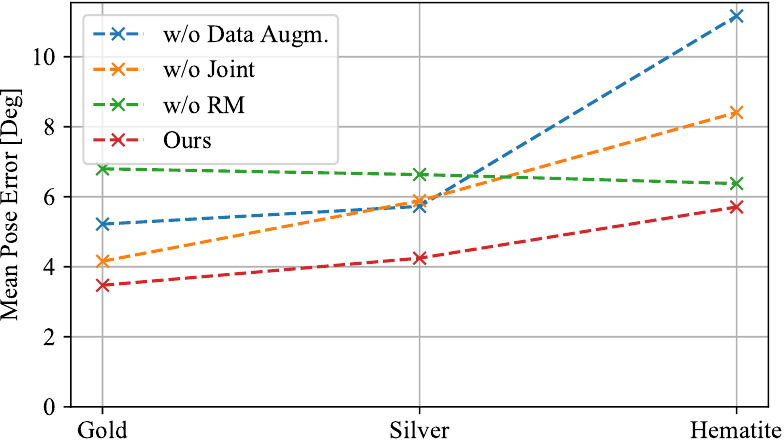}
    }
    \\
    \subfloat[][Accuracy for each shape]{
        \includegraphics[keepaspectratio, width=0.45\linewidth]{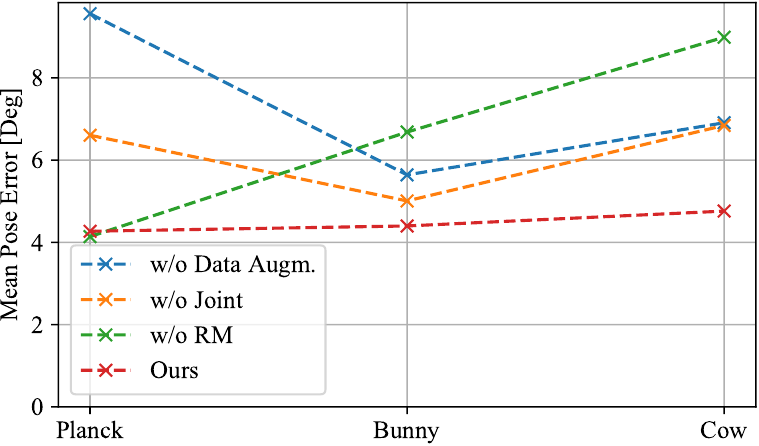}
    }
    \caption{
        Mean camera pose (relative rotation) estimation error for each illumination, each material, and each shape in the synthetic test data. The results show the effectiveness and the robustness of our full method. Please see \cref{fig:synth-test-assets} for the names of the illuminations, the materials, and the shapes.
    }
    \label{fig:synth-ablation-details}
\end{figure*}

\Cref{fig:synth-ablation-details} shows the mean camera pose (relative rotation) estimation error for each illumination, each material, and each shape in the synthetic test data. Our results are at least comparable to and in most cases more accurate than those of other methods. These results show the effectiveness and robustness of our full method.

\section{Evaluation on Real Data}

\begin{figure}[p]
  \centering
  \includegraphics[keepaspectratio, width=\linewidth]{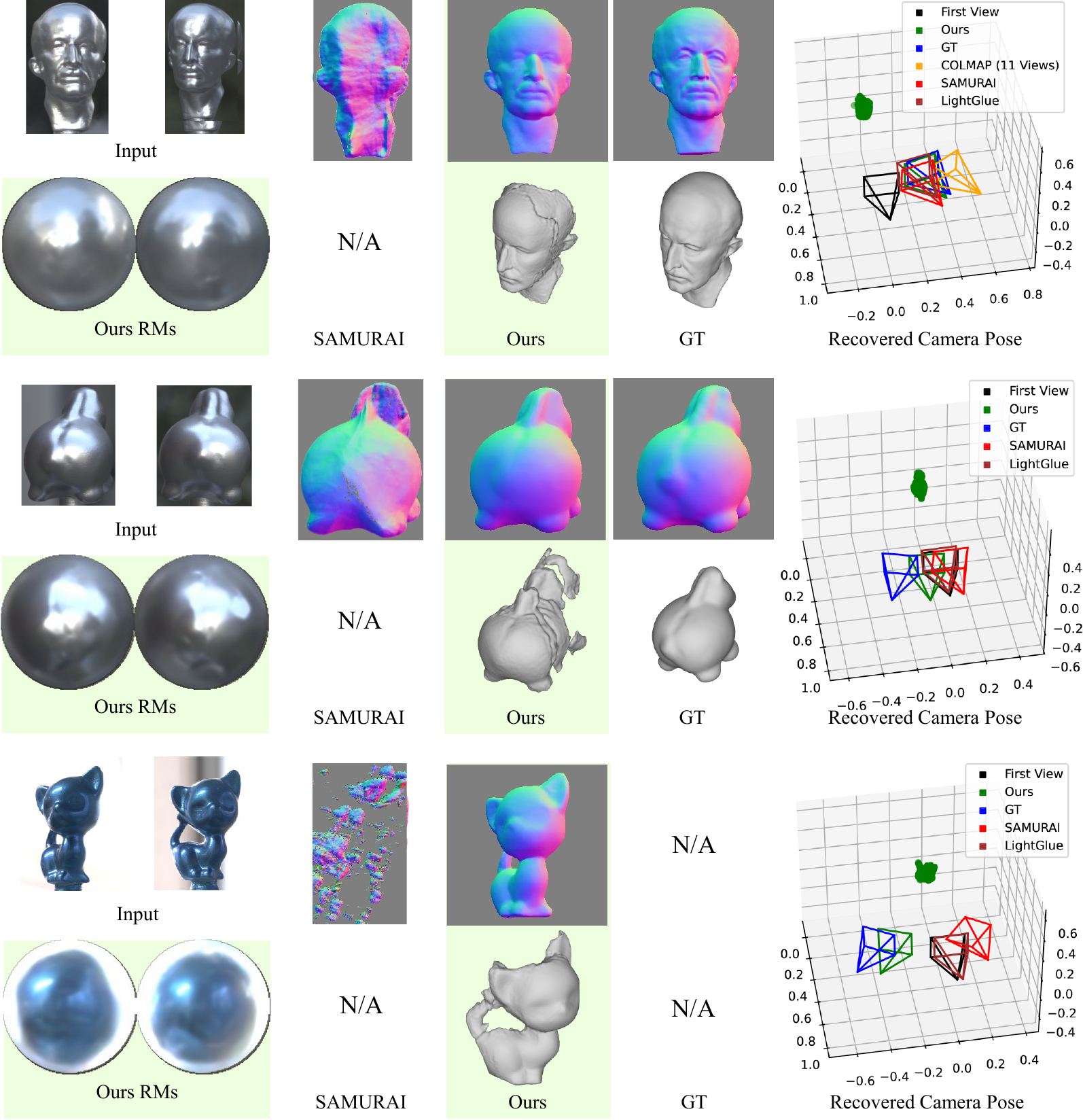}
  \caption{Qualitative results on real image pairs from the nLMVS-Real dataset~\cite{yamashita2023nlmvs} and on an image pair captured and calibrated by ourselves (the third object).}
  \label{fig:result_real_supp}
\end{figure}

\begin{figure}[t]
  \centering
  \includegraphics[keepaspectratio, width=\linewidth]{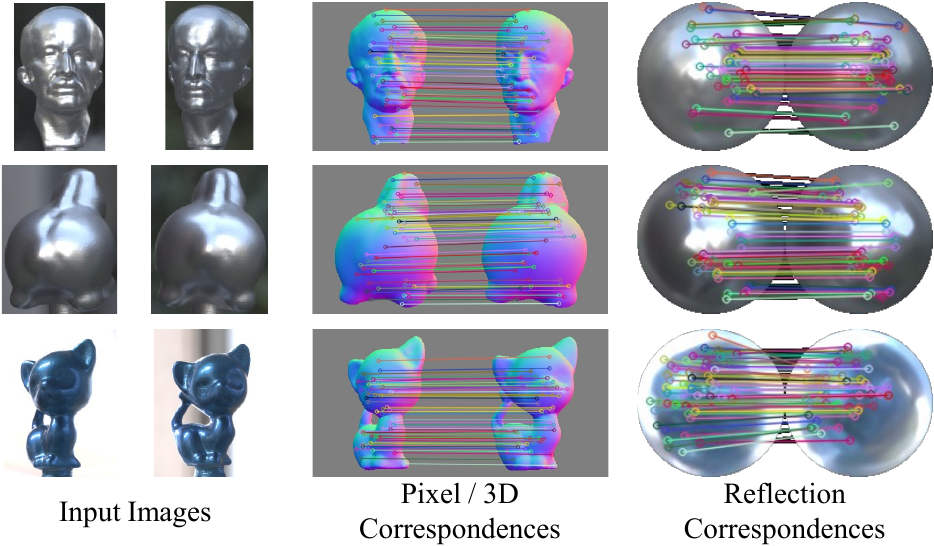}
  \caption{Correspondences detected from the real image pairs in the nLMVS-Real dataset~\cite{yamashita2023nlmvs} and on an image pair captured by ourselves.}
  \label{fig:result_corrs_real}
\end{figure}

\Cref{fig:result_real_supp,fig:result_corrs_real} show qualitative results on real image pairs from the nLMVS-Real dataset~\cite{yamashita2023nlmvs} and one captured by ourselves. Our method successfully recovers plausible camera poses, surface geometry, surface normals, reflectance maps, and correspondences across views even for real objects without requiring textures and overlapping background.

\subsubsection{Comparison with DUSt3R~\cite{wang24dust3r}}
\begin{figure}[t]
  \centering
  \includegraphics[keepaspectratio, width=0.9\linewidth]{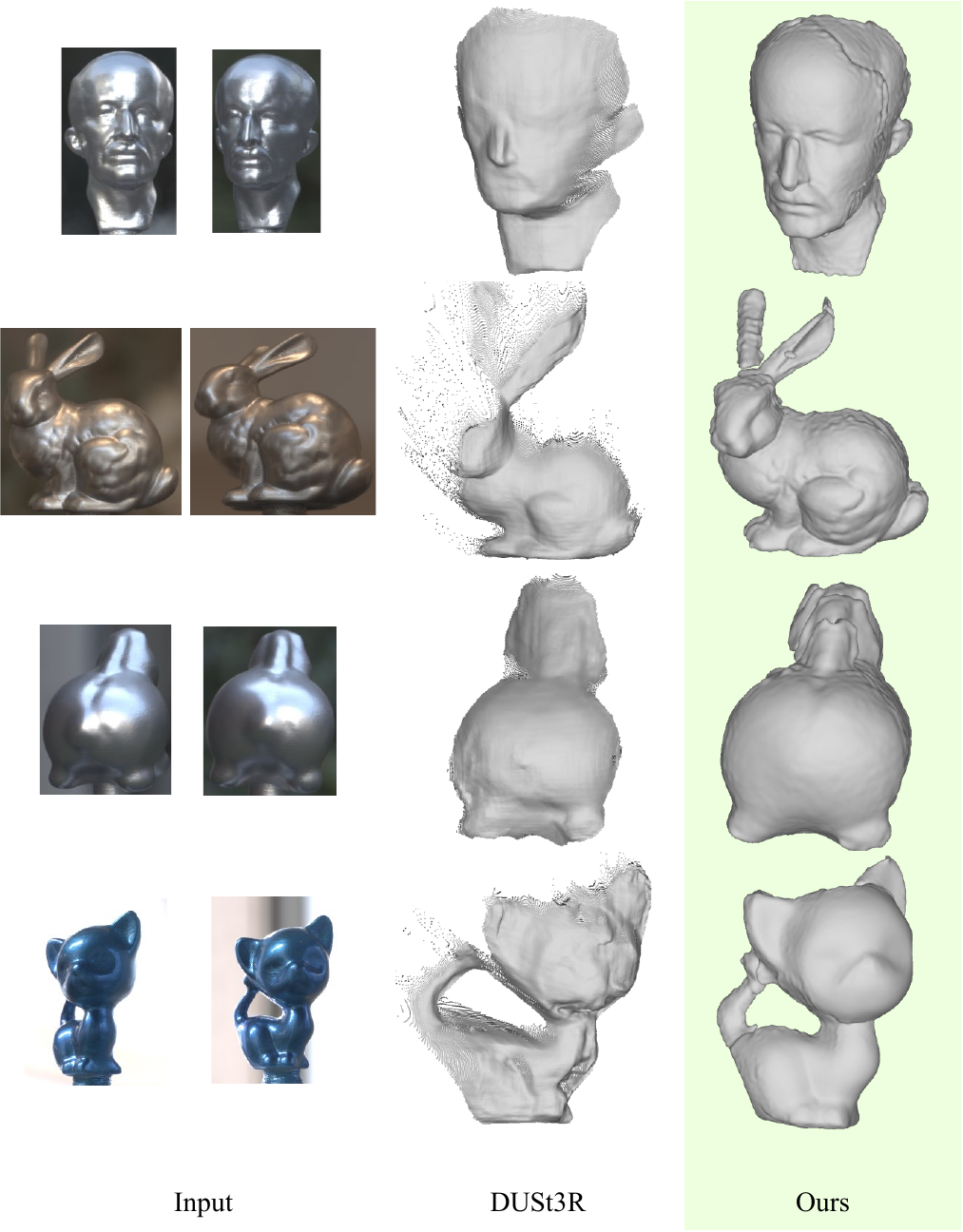}
  \caption{Qualitative comparison with DUSt3R~\cite{wang24dust3r}. The geometry recovered with DUSt3R is coarse. In contrast, our method successfully recovers details of the surface (e.g., bumps of the Bunny).}
  \label{fig:comparison-dust3r}
\end{figure}

We also compare our method with DUSt3R. As the output of DUSt3R is a point map and relative camera pose estimation from it does not resolve the overall scale, we can only achieve qualitative comparison. \Cref{fig:comparison-dust3r} shows the results. The geometry recovered with DUSt3R is coarse. In contrast, our method successfully recovers details of the surface (e.g., bumps of the Bunny).

\subsection{Scenes with Textured and Textureless Objects}

\begin{figure}[t]
  \centering
  \includegraphics[keepaspectratio, width=\linewidth]{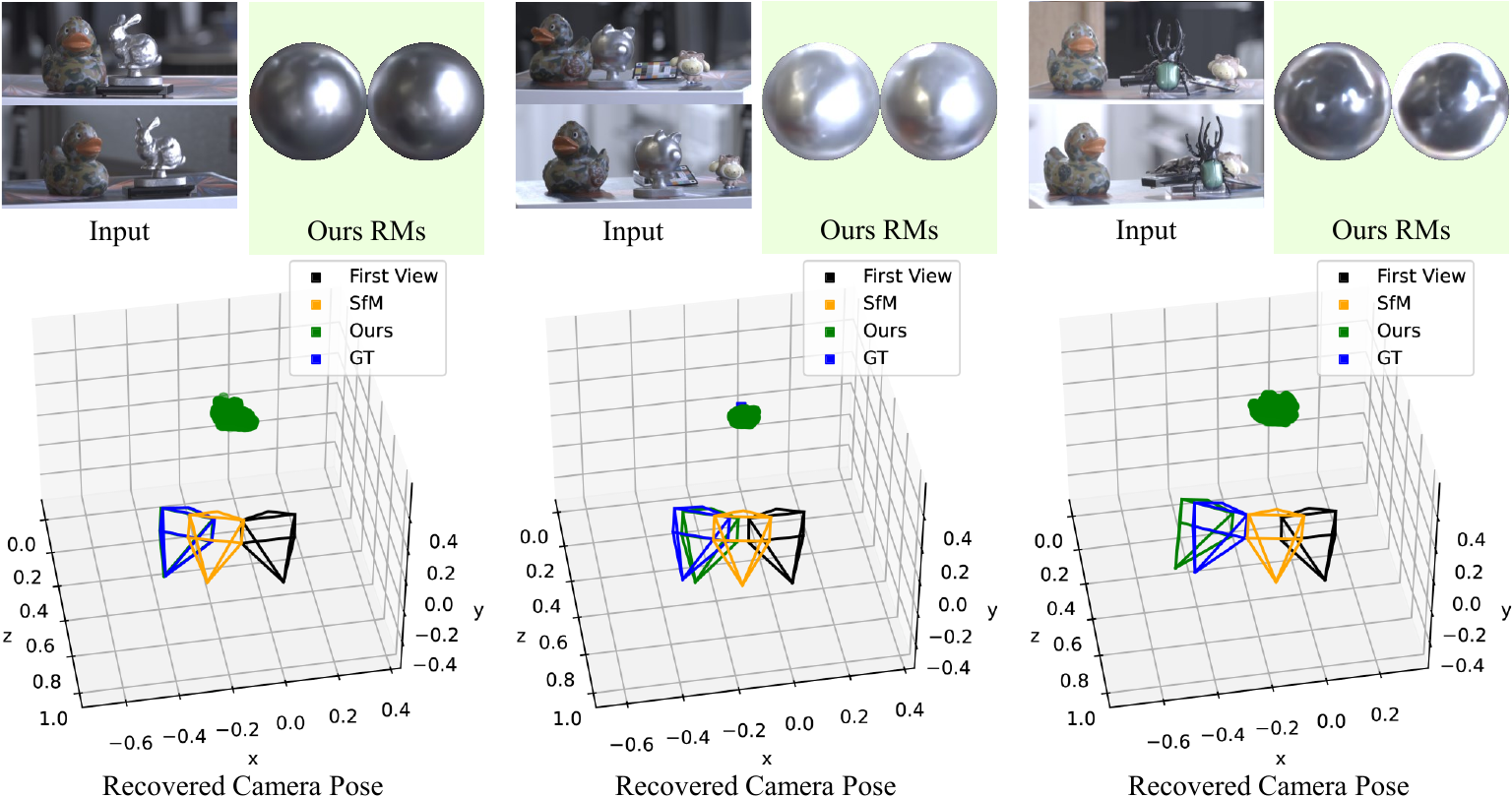}
  \caption{Reflectance maps and camera poses recovered from two-view real images of scenes with textured and textureless, non-Lambertian objects. Leveraging different types of correspondences from different objects in the same scene leads to more accurate camera pose estimation.}
  \label{fig:result_scene_estimation_supp}
\end{figure}

\begin{figure}[t]
  \centering
  \includegraphics[keepaspectratio, width=\linewidth]{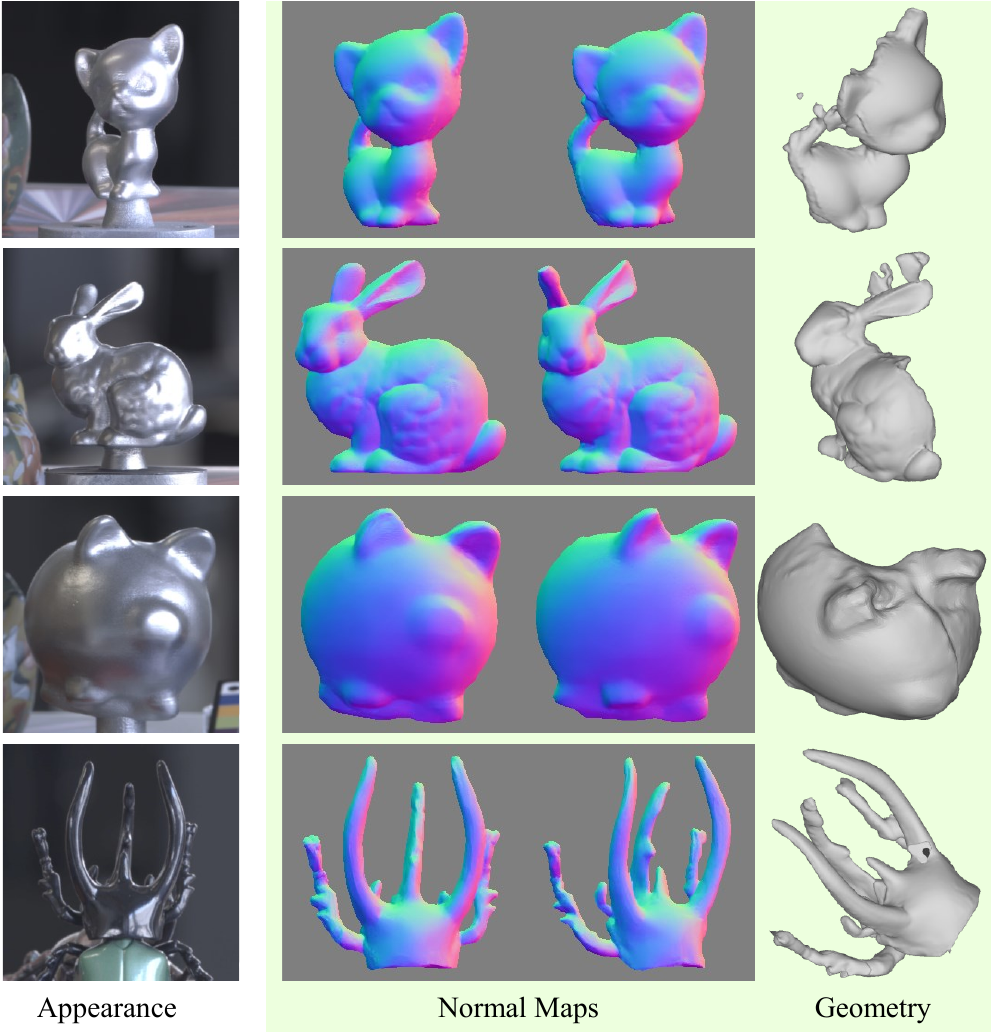}
  \caption{Normal maps and surface geometry of textureless, non-Lambertian objects recovered from the two-view real images of scenes with different objects. The recovered geometry is qualitatively plausible for the captured objects.}
  \label{fig:result_scene_estimation_geometry}
\end{figure}

\begin{figure}[t]
  \centering
  \includegraphics[keepaspectratio, width=\linewidth]{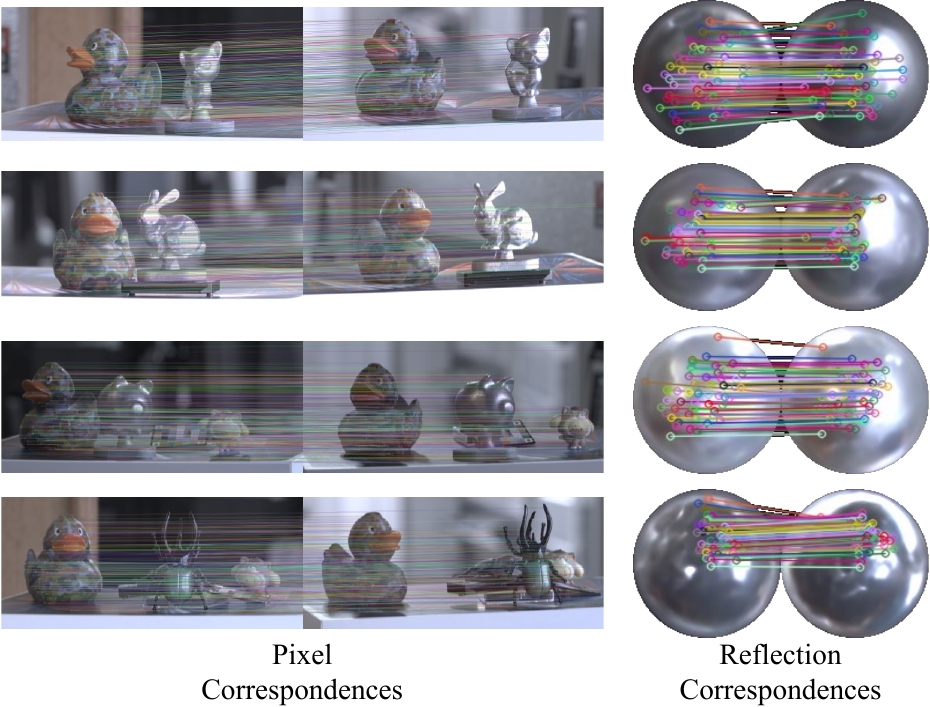}
  \caption{Correspondences detected from the real images of scenes with textured and textureless, non-Lambertian objects.}
  \label{fig:result_corrs_real_scene}
\end{figure}

\Cref{fig:result_scene_estimation_supp,fig:result_scene_estimation_geometry,fig:result_corrs_real_scene} show camera poses, reflectance maps, normal maps, surface geometry, and correspondences estimated from two-view real images of scenes with textured and textureless, non-Lambertian objects. By leveraging different types of correspondences from different objects in the same scene, we can recover accurate camera poses and plausible surface geometry for complex in-the-wild scenes.

\afterpage{\clearpage}
\newpage

\section{Ambiguity in The Estimation with Conventional Correspondences}
\label{sec:ambiguity_proof}

In this section, we prove that, under orthographic projection (\eg, cameras are distant from the object), relative camera pose estimation between two views suffers from a fundamental ambiguity that cannot be resolved even when correct pixel and 3D correspondences are provided.

Harris~\cite{harris91orthographicsfm} showed that a least-square objective for establishing pixel correspondences (Eq. (14) in the main text) is equivalent to
\begin{equation}
    \min_{z_2^i, t_x, t_y}f_\IM' = \min_{z_2^i, t_x, t_y}\sum_i f_u^i + f_v^i\,,
    \label{eq:obj_pixel_original}
\end{equation}
where  $z_2^i$ is the depth of a surface point in the second view for the $i$-th correspondence, $t_x$ and $t_y$ are the $x$ and $y$ elements of the relative translation vector, respectively, $f_u^i$ and $f_v^i$ are reprojection errors along the x and y axes, respectively, 
\begin{equation}
    f_u^i = (r_{11}u_2^i+r_{12}v_2^i+ r_{13}z_2^i+t_x-u_1^i)^2\,,
\end{equation}
\begin{equation}
    f_v^i = (r_{21}u_2^i+r_{22}v_2^i+ r_{23}z_2^i+t_y-v_1^i)^2\,,
\end{equation}
and $r_{ij}$ is the $(i,j)$ element of the relative rotation $\mathrm{R_{21}}$. Please see Harris~\cite{harris91orthographicsfm} for details. 

Let us derive the relationship between the unresolvable parameter $\eta$ and corresponding depth estimates (surface geometry) ${z_2^i}$. As the depth values that minimize \cref{eq:obj_pixel_original} should satisfy
\begin{equation}
    \frac{\partial}{\partial z_2^i} f_\IM' = 0\,,
\end{equation}
we have
\begin{equation}
    z_2^i = -\frac{r_{13}(\tilde{u}^i+t_x)+r_{23}(\tilde{v}^i+t_y)}{r_{13}^2+r_{23}^2}\,,
    \label{eq:optimal_depth}
\end{equation}
where 
\begin{equation}
    \tilde{u}^i = r_{11}u_2^i + r_{12}v_2^i - u_1^i\,,
\end{equation}
\begin{equation}
    \tilde{v}^i = r_{21}u_2^i + r_{22}v_2^i - v_1^i\,.
\end{equation}
The elements of $\mathrm{R_{21}}$ are related to their corresponding Euler angles $(\phi, \eta, \theta)$ by
\begin{align}
    r_{11} &= \cos\phi\cos\theta + \sin\phi\sin\theta\cos\eta \,, \\
    r_{12} &= \cos\phi\sin\theta - \sin\phi\cos\theta\cos\eta \,, \\
    r_{13} &= \sin\phi \sin\eta \,, \\
    r_{21} &= \sin\phi\cos\theta - \cos\phi\sin\theta\cos\eta \,, \\
    r_{22} &= \sin\phi\sin\theta + \cos\phi\cos\theta\cos\eta \,, \\
    r_{23} &= -\cos\phi \sin\eta \,, \\
    r_{31} &= -\sin\theta \sin\eta \,, \\
    r_{32} &= \cos\theta \sin\eta \,, \\
    r_{33} &= \cos\eta \,.
\end{align}
By substituting these for \cref{eq:optimal_depth}, we have
\begin{equation}
    z_2^i = \frac{1}{\tan\eta} n_\theta^i - \frac{1}{\sin\eta} (n_\phi^i - n_{t,\phi})\,,
    \label{eq:optimal_depth_euler}
\end{equation}
where
\begin{equation}
    n_\theta^i = u_2^i\sin\theta  - v_2^i\cos\theta \,,
\end{equation}
\begin{equation}
    n_\phi^i = u_1^i\sin\phi  - v_1^i\cos\phi \,,
\end{equation}
\begin{equation}
    n_{t,\phi} = t_x\sin\phi  - t_y\cos\phi \,.
\end{equation}

Let us assume that $\hat{\eta}$ is an inaccurate estimate of the unresolvable Euler angle and $\eta$ is the corresponding ground truth. These two quantities $\hat{\eta}$ and $\eta$ are tied by
\begin{equation}
    \frac{1}{\sin\hat{\eta}} = \frac{\sin\eta}{\sin\hat{\eta}}\frac{1}{\sin\eta}\,,
    \label{eq:inv_sin_eta}
\end{equation}
\begin{equation}
    \frac{1}{\tan\hat{\eta}} = \frac{\sin\eta}{\sin\hat{\eta}}\frac{1}{\tan\eta} + \left(\frac{\cos\hat{\eta}}{\cos\eta} - 1\right) \frac{\sin\eta}{\sin\hat{\eta}}\frac{1}{\tan\eta}\,.
    \label{eq:inv_tan_eta}
\end{equation}
By using \cref{eq:optimal_depth_euler,eq:inv_sin_eta,eq:inv_tan_eta}, we can relate (inaccurate) depth estimates $\hat{z}_2^i$ that correspond to $\hat{\eta}$ with the ground truth $z_2^i$
\begin{equation}
    \hat{z}_2^i = \lambda_{\hat{\eta}} z_2^i + \mu_{\hat{\eta}} u_2^i + \nu_{\hat{\eta}} v_2^i + c_{\hat{\eta}} \,,
\end{equation}
where
\begin{equation}
    \lambda_{\hat{\eta}} = \frac{\sin\eta}{\sin\hat\eta}\,,
\end{equation}
\begin{equation}
    \mu_{\hat{\eta}} = \left(\frac{\cos\hat{\eta}}{\cos\eta} - 1\right) \frac{\sin\eta}{\sin\hat{\eta}}\frac{1}{\tan\eta} \sin\theta \,,
\end{equation}
\begin{equation}
    \nu_{\hat{\eta}} = -\left(\frac{\cos\hat{\eta}}{\cos\eta} - 1\right) \frac{\sin\eta}{\sin\hat{\eta}}\frac{1}{\tan\eta} \cos\theta \,,
\end{equation}
\begin{equation}
    c_{\hat{\eta}} = \frac{1}{\sin\hat{\eta}} \left(\hat{n}_{t,\phi} - n_{t,\phi}\right)\,,
\end{equation}
and $\hat{n}_{t,\phi}$ is an estimate of $n_{t,\phi}$ that corresponds to $\hat{\eta}$. This relationship is identical to the generalized bas-relief (GBR) transformation~\cite{belhumeur99basrelief}
\begin{equation}
    d'[u_2^i, v_2^i] = \lambda_2 d[u_2^i, v_2^i] + \mu_2 u_2^i + \nu_2 v_2^i\,, 
\end{equation}
except for the global offset $c_{\hat{\eta}}$. Therefore, for any possible $\hat{\eta}$ and corresponding inaccurate surface geometry, there is a GBR transformation
\begin{equation}
    \hat{\mathrm{G}}_2 = \mathrm{G}_2 \mathrm{G}_{\hat{\eta},2}^{-1} \,,
\end{equation}
that transforms surface normals of the inaccurate surface geometry to those of the recovered normal maps in the second view. $\mathrm{G}_2$ is the transformation from the ground-truth geometry to the recovered normal map and $\mathrm{G}_{\hat{\eta},2}$ is that from the ground truth to the inaccurate geometry
\begin{equation}
    \mathrm{G}_{\hat{\eta},2} = 
    \begin{pmatrix}
        1 & 0 & o \\
        0 & 1 & 0 \\
        \mu_{\hat{\eta}} & \nu_{\hat{\eta}} & \lambda_{\hat{\eta}}
    \end{pmatrix}
    \,.
\end{equation}
Similarly, we can also derive the relationship between $\hat{\eta}$ and the GBR transformation of the first view $\hat{\mathrm{G}}_1$. Note that, when the signs of $\eta$ and $\hat{\eta}$ are different, $ \lambda_{\hat{\eta}}$ becomes negative, \ie, the surface geometry is inverted. This causes corresponding estimates of $\lambda_1$ and $\lambda_2$ to be negative. We can easily exclude such a solution in practice.

\section{Analytical Decomposition of $\mathrm{G}_{21}$}
\label{sec:analytical_decomposition}
As we discussed in the main text and in \cref{sec:ambiguity_proof}, given just the pixel and 3D correspondences, we can only obtain unique solutions for the combined transformation $\mathrm{G}_{21} \equiv \mathrm{G_1}^{-\transp}\mathrm{R_{21}}\mathrm{G_2}^\transp$ and the two Euler angles $(\theta, \phi)$. However, if the remaining Euler angle $\eta$ is given in addition to $\theta$ and $\phi$, \ie, if $\mathrm{R_{21}}$ is determined, we can solve for the parameters of the generalized bas-relief transformations analytically. For this, we derive a two-step relative rotation estimation algorithm as detailed in \cref{sec:algorithm_details}. 

From the definitions of $\mathrm{G_{21}}$, $\mathrm{G_1}$, $\mathrm{G_2}$, and $\mathrm{R_{21}}$, we have
\begin{equation}
    \mathrm{G_{21}} =
    \begin{pmatrix}
        \gamma_{11} & \gamma_{12} & \gamma_{13} \\
        \gamma_{21} & \gamma_{22} & \gamma_{23} \\
        \gamma_{31} & \gamma_{32} & \gamma_{33}
    \end{pmatrix}
    \,,
\end{equation}
\begin{align}
    \gamma_{11} &= r_{11} - r_{31} \frac{\mu_1}{\lambda_1}\,, \label{eq:m11} \\
    \gamma_{12} &= r_{12} - r_{32} \frac{\mu_1}{\lambda_1}\,, \label{eq:m12} \\
    \gamma_{13} &= r_{11}\mu_2 + r_{12}\nu_2 + r_{13}\lambda_2 -\mu_1 \gamma_{33} \label{eq:m13} \,, \\
    \gamma_{21} &= r_{21} - r_{31} \frac{\nu_1}{\lambda_1}\,, \label{eq:m21} \\
    \gamma_{22} &= r_{22} - r_{32} \frac{\nu_1}{\lambda_1}\,, \label{eq:m22} \\
    \gamma_{23} &= r_{21}\mu_2 + r_{22}\nu_2 + r_{23}\lambda_2 -\nu_1 \gamma_{33} \label{eq:m23} \,, \\
    \gamma_{31} &= \frac{r_{31}}{\lambda_1}\,, \label{eq:m31} \\
    \gamma_{32} &= \frac{r_{32}}{\lambda_1}\,, \label{eq:m32} \\
    \gamma_{33} &= \frac{1}{\lambda_1}(r_{31}\mu_2 + r_{32}\nu_2 + r_{33}\lambda_2)\,, \label{eq:m33}
\end{align}
where $r_{ij}$ is the $(i,j)$ element of $\mathrm{R_{21}}$. From \cref{eq:m31} and \cref{eq:m32}, we have
\begin{equation}
    \lambda_1^2 = \frac{r_{31}^2+r_{32}^2}{\gamma_{31}^2+\gamma_{32}^2}\,.
\end{equation}
Since $\lambda_1$ is positive, by taking the square root of both sides, we have a solution for $\lambda_1$
\begin{equation}
    \lambda_1 = \sqrt{\frac{r_{31}^2+r_{32}^2}{\gamma_{31}^2+\gamma_{32}^2}}\,.
\end{equation}
As there is a relationship regarding determinants of the matrices
\begin{equation}
    \det \left(\mathrm{G_{21}}\right) = \det \left(\mathrm{G_{1}}^\mathrm{-T}\right)\det \left(\mathrm{R_{21}}\right)\det \left(\mathrm{G_{2}}^\mathrm{T}\right)=\frac{\lambda_2}{\lambda_1}\,,
\end{equation}
we also have a solution for $\lambda_2$:
\begin{equation}
    \lambda_2 = \det \left(\mathrm{\mathrm{G_{21}}}\right) \lambda_1\,.
\end{equation}
As \cref{eq:m11,eq:m12,eq:m13,eq:m21,eq:m22,eq:m23,eq:m33} are linear equations for the remaining unknown parameters, we can analytically solve these equations in the least-squares sense. Note that, we empirically found that there is a solution for these equations only when the sign of the given $\eta$ and one of the ground truth are identical. This is consistent with the discussion in \cref{sec:ambiguity_proof}.

\section{Two-Step Relative Rotation Estimation}
\label{sec:algorithm_details}

\begin{figure}[!t]
\begin{algorithm}[H]
    \scriptsize
    \algsetup{linenosize=\scriptsize}
    \SetAlCapFnt{\small}
    \SetAlCapNameFnt{\small}
    \caption{Relative Rotation Estimation}
    \label{alg:two-step}
    \begin{algorithmic}[1]
        \STATE // Estimation of $\mathrm{G}_{21}$ and $(\phi, \theta)$
        \STATE // with pixel and 3D correspondences
        \STATE $bestScore1 \leftarrow 0$
        \FOR{$i \leftarrow 1$ to $L_1$}
            \STATE $S_1 \leftarrow$ random $M_1$ indices from $N_\NM$ indices
            \STATE $\hat{\mathrm{G}}_{21}, \hat{\phi}, \hat{\theta} \leftarrow \argmin_{\mathrm{G}_{21}} f_1(S_1)$
            \IF{$g_1\left(\hat{\mathrm{G}}_{21}, \hat{\phi}, \hat{\theta}\right) > bestScore1$ }
                \STATE $bestScore1 \leftarrow g_1\left(\hat{\mathrm{G}}_{21}, \hat{\phi}, \hat{\theta}\right)$
                \STATE $bestModel1 \leftarrow \hat{\mathrm{G}}_{21}, \hat{\phi}, \hat{\theta}$
            \ENDIF
        \ENDFOR
        \STATE $\hat{\mathrm{G}}_{21}, \hat{\phi}, \hat{\theta} \leftarrow bestModel1$
        \STATE // Decomposition of $\mathrm{G}_{21}$
        \STATE // with $g_2(\cdot)$ which exploits reflection correspondences
        \STATE $bestScore2 \leftarrow 0$
        \FOR{$\eta \leftarrow$ 1 deg to 359 deg}
            \STATE $\hat{\mathrm{R}}_{21},\hat{\mathrm{G}}_{1},\hat{\mathrm{G}}_{2} \leftarrow$ decomposition($\hat{\mathrm{G}}_{21}; \hat{\phi}, \eta, \hat{\theta}$)
            \IF{$g_2\left(\hat{\mathrm{R}}_{21},\hat{\mathrm{G}}_{1},\hat{\mathrm{G}}_{2}\right) > bestScore2$}
                \STATE $bestScore2 \leftarrow g_2\left(\hat{\mathrm{R}}_{21},\hat{\mathrm{G}}_{1},\hat{\mathrm{G}}_{2}\right)$
                \STATE $bestModel2 \leftarrow \hat{\mathrm{R}}_{21},\hat{\mathrm{G}}_{1},\hat{\mathrm{G}}_{2}$
            \ENDIF
        \ENDFOR
        \RETURN $bestModel2$

    \end{algorithmic}
\end{algorithm}
\end{figure}

\Cref{alg:two-step} shows a pseudo code of the two-step relative rotation estimation from the three types of correspondences. In the first step, using $N_\NM$ pixel and $N_\NM$ 3D correspondences, we obtain estimates of the combined transformation $\mathrm{G}_{21}$ and the two Euler angles $(\phi, \theta)$ based on RANSAC~\cite{fischler81ransac}. For each of the $L_1$ iterations, we first build a set of randomly sampled $M_1$ indices for the pixel and 3D correspondences. We then obtain estimates of the combined transformation $\hat{\mathrm{G}}_{21}$ and the Euler angles $(\hat{\phi}, \hat{\theta})$ by minimizing an objective function
\begin{equation}
    f_1(S_1) = f_\IM(S_1) + f_\NM^{(12)}(S_1) + f_\NM^{(21)}(S_1)\,,
\end{equation}
with respect to the relative rotation and parameters of the generalized bas-relief transformations ($\mu_1, \nu_1, \lambda_1, \mu_2, \nu_2, \lambda_2$). $f_\IM(S_1)$ and $f_\NM^{(jk)}(S_1)$ are similar to those in Eq.~(13) in the main text. We compute them using only the sampled correspondences
\begin{equation}
    f_\IM(S_1) = \frac{1}{{N_\NM}}\sum_{i\in S_1} \left( t_\phi^i - t_\theta^i \right)^2\,,
\end{equation}
\begin{equation}
    f_\NM^{(jk)}(S_1) = \frac{1}{{N_\NM}}\sum_{i \in S_1} \left\|\bN_j[u_j^i, v_j^i]  - \Norm(\mathrm{G}_{kj} \bN_k[u_k^i, v_k^i])\right\|^2\,,
\end{equation}
with
\begin{equation}
    t_\phi^i = u_1^i \cos\phi + v_1^i \sin \phi\,,
\end{equation}
\begin{equation}
    t_\theta^i = u_2^i \cos\theta + v_2^{(i)} \sin \theta\,.
\end{equation}
In practice, we achieve this minimization by using an off-the-shelf solver for nonlinear optimization~\cite{2020SciPy-NMeth,branch99trf}. We evaluate the goodness of the estimates using a function
\begin{equation}
    g_1\left({\mathrm{G}}_{21}, \phi, \theta\right) = \frac{1}{N_\NM}\sum_{i=1}^{N_\NM} \exp\left(-\frac{p_i^2}{T_\IM^2}\right) \exp\left(-\frac{\left(q_i^{(12)}\right)^2+\left(q_i^{(21)}\right)^2}{T_\NM^2}\right) \,,
\end{equation}
where $p_i$ and $q_i^{(jk)}$ are residuals with respect to the pixel and the 3D correspondences
\begin{equation}
    p_i = t_\phi^i - t_\theta^i \,,
    \label{eq:pixel_reproj_error}
\end{equation}
\begin{equation}
    q_i^{(jk)} = \arccos\left(\bN_j[u_j^i, v_j^i]  \cdot \Norm(\mathrm{G}_{kj} \bN_k[u_k^i, v_k^i])\right) \,,
\end{equation}
and $T_\IM$ and $T_\NM$ are corresponding thresholds.
This function computes the approximated number of correspondences that are consistent with the obtained estimates. We select a set of the estimates $(\hat{\mathrm{G}}_{21}, \hat{\phi}, \hat{\theta})$ that maximizes this function.

In the second step, we discretize $\eta$ in 360 angles and, for each $\eta$, decompose the estimate of the combined transformation $\hat{\mathrm{G}}_{21}$ into the relative rotation $\mathrm{R}_{21}$ and the GBR transformations $\hat{\mathrm{G}}_1$ and $\hat{\mathrm{G}}_2$ by using the method in \cref{sec:analytical_decomposition}. We then select a set of the estimates that maximizes
\begin{equation}
    g_2\left(\mathrm{R}_{21}, \mathrm{G}_1, \mathrm{G}_2\right) = s_\mathrm{reg} \frac{1}{N_\RM}\sum_{i=1}^{N_\RM} \exp\left(-\frac{\left(r_i^{(12)}\right)^2+\left(r_i^{(21)}\right)^2}{T_\RM^2}\right) \,,
\end{equation}
where $r_i^{(jk)}$ is the residual with respect to the reflection correspondences
\begin{equation}
    r_i^{(jk)} = \arccos\left( \bN_j[u_j^i, v_j^i] \cdot
    \Omega^{(jk)}( \bN_k[u_k^i, v_k^i] ) \right) \,,
\end{equation}
and $T_\RM$ is a corresponding threshold.

In practice, we set $L_1$, $T_\IM$, $T_\NM$, and $T_\RM$ to be 200, 5 pixels, 20 degrees, and 10 degrees, respectively. We set $M_1$ to be 4 according to empirical evaluation below.

\subsection{Empirical Evaluation about The Number of Correspondences}
\label{sec:exp_num_corrs}

As mentioned in Section~3.4 in the main text, we conduct empirical evaluation of the number of correspondences required for each step of the two-step algorithm. We created 10000 sets of synthetic correspondences by randomly sampling surface normal orientations, relative rotation matrices, and parameters of the GBR transformation. For each set of correspondences, we optimized all the unknown parameters with all available correspondences (\ie, Eq.~(13) in the main text) using an off-the-shelf solver~\cite{2020SciPy-NMeth,branch99trf} and checked if the estimates are accurate. 

\begin{wrapfigure}[13]{r}{0.5\textwidth}
  \centering
  \includegraphics[keepaspectratio, width=\linewidth]{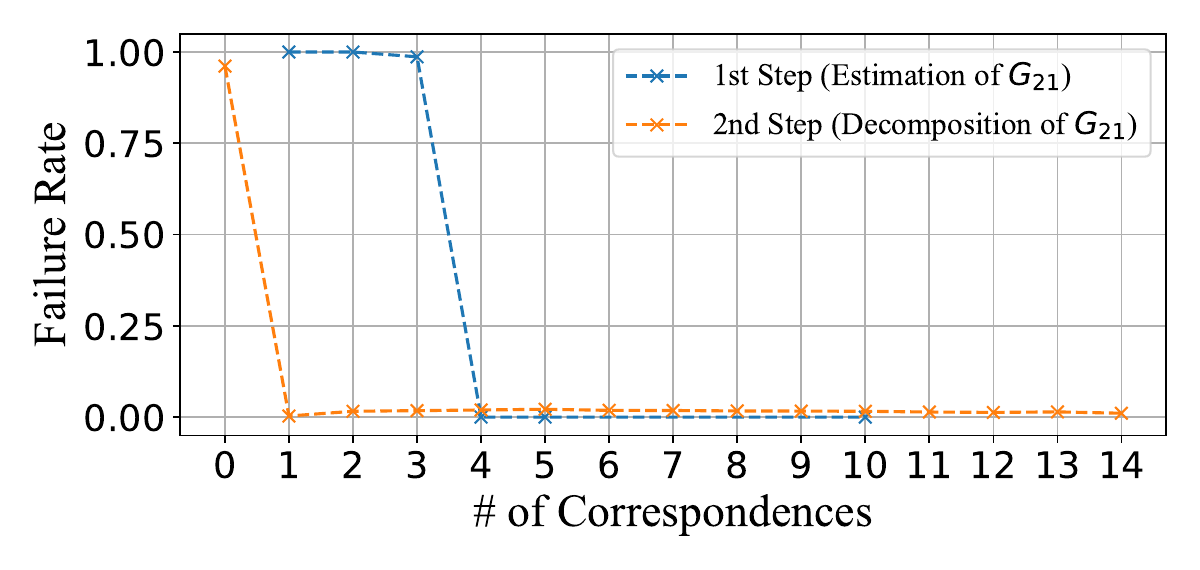}
  \caption{Empirical analysis of the number of correspondences required for optimization of the camera pose and the two GBR transforms of two images. See text for details.}
  \label{fig:num-corrs}
\end{wrapfigure}

The blue line in \cref{fig:num-corrs} shows the relationship between the number of correspondences in normal maps (pixel and 3D correspondences) and rate of failure cases where the error (Frobenius norm) between estimated combined transformation $\mathrm{G_{21}}$ and the ground truth is higher than 0.01. 
The results show that four correspondences are required for the estimation of $\mathrm{G_{21}}$. The orange line in \cref{fig:num-corrs} shows the relationship between the number of reflection correspondences and rate of failure cases where the error between the estimated camera pose and the ground truth is higher than 0.1 degree. In this experiment, we used four pixel and four 3D correspondences in addition to the reflection correspondences. One reflection correspondence resolves the remaining ambiguity. Note that, the leftmost point in the orange line shows that, without reflection correspondences, we cannot decompose the combined transformation $\mathrm{G_{21}}$. 
We hope to derive a theoretical justification of these results in our future work.

\section{Single-View Estimation with DeepShaRM~\cite{yamashita2023deepsharm}}
As explained in the main paper, given posed multi-view images and an initial coarse estimate of surface geometry (\eg, a visual hull or a bounding sphere) represented by a 3D grid of a signed distance function, DeepShaRM~\cite{yamashita2023deepsharm} can jointly recover geometry, reflectance maps, and normal maps by alternating between (1) learning-based reflectance map estimation for each view from an image and the geometry estimate, (2) learning-based surface normal estimation for each view from an image and the estimated reflectance map, and (3) multi-view geometry optimization with the estimated surface normals. Note that the learning-based reflectance map estimation and the learning-based surface normal estimation are trained using synthetic images in a supervised manner.

While DeepShaRM~\cite{yamashita2023deepsharm} is originally designed for posed multi-view images, we found that its joint shape and reflectance map estimation framework can work even with a single-view image. To reduce the computational cost, instead of using a 3D grid of a signed distance function, we represent the surface geometry as a 2D grid of the surface depths. As the single-view estimate suffers from the bas-relief ambiguity, during the geometry optimization with the estimated surface normals, we impose a regularization loss
\begin{equation}
    L_\mathrm{reg} = \left\{\begin{pmatrix}0 \\ 0 \\ l\end{pmatrix} -\frac{1}{|P|}\sum_{u,v\in P}\mathbf{N}[u,v]\right\}^2\,,
\end{equation}
where $\mathbf{N}[u,v]$ is a normal map extracted from the depth grid, $P$ is a set of pixels inside the object, and $|P|$ is the number of such pixels. As we can assume that the surface is roughly fronto-parallel, we enforce the x and y components of the average surface normal orientation to be zero. We also encourage its z component to be identical to its average value in training data $l$. As the value of the z component corresponds to the flatness of the surface (\ie, higher values correspond to almost all surface normals facing the camera), this loss ensures that the surface is not extremely flat or sharp. For the remaining estimation, we follow the method described in \cite{yamashita2023deepsharm}.

\section{Learning-Based Correspondence Detection}

\begin{figure}[t]
  \centering
  \includegraphics[keepaspectratio, width=\linewidth]{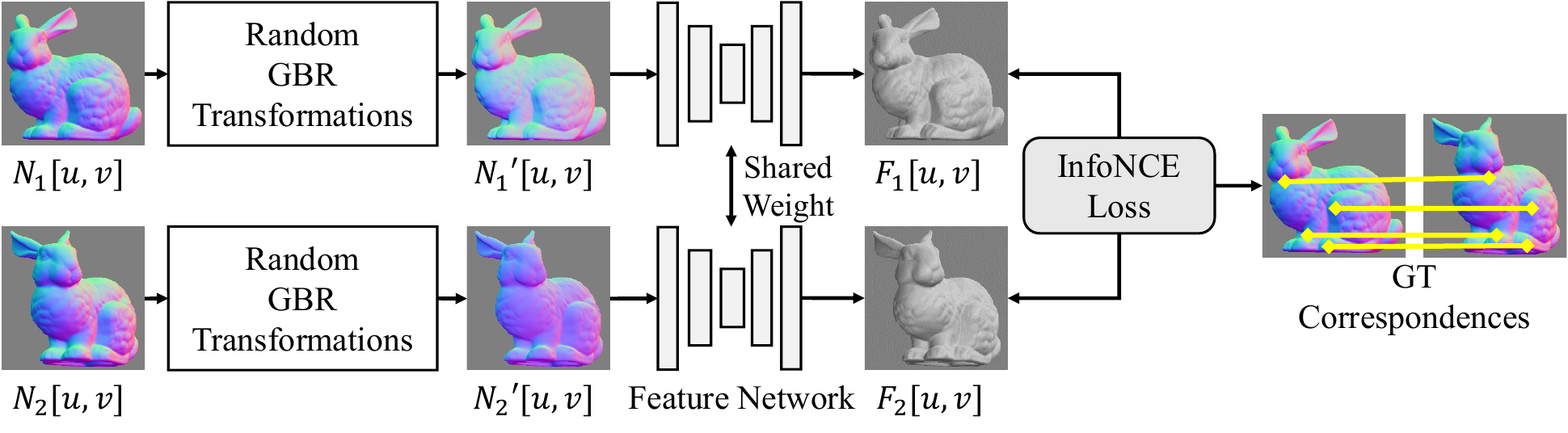}
  \caption{We train a deep neural network to extract view-independent, per-pixel features for correspondence matching. Contrastive learning using InfoNCE loss~\cite{oord18infonce} with data augmentation with the GBR transform enables feature extraction for distortion-robust correspondence matching.}
  \label{fig:fea-ext-training}
\end{figure}

\Cref{fig:fea-ext-training} depicts how we train a feature extraction network for correspondence detection from normal maps. As explained in Section~3.5 in the main text, using synthetic normal maps and corresponding ground truth matches as training data, we train the network by contrastive learning. We also augment the input normal maps by generalized bas-relief (GBR) transformations so that the network can learn to extract features robust to the estimation errors caused by the bas-relief ambiguity.

\subsection{Network Architecture}
We use the UNet-based architecture described in~\cite{yamashita2023deepsharm} (Tab. 5) to implement the feature extraction networks for correspondence detection. Each of the feature extraction networks consists of a series of two UNets. We set the number of output channels of the UNets to 36.

\section{Estimation for Scenes with Textured and Textureless Objects}

\begin{figure}[t]
  \centering
  \includegraphics[keepaspectratio, width=\linewidth]{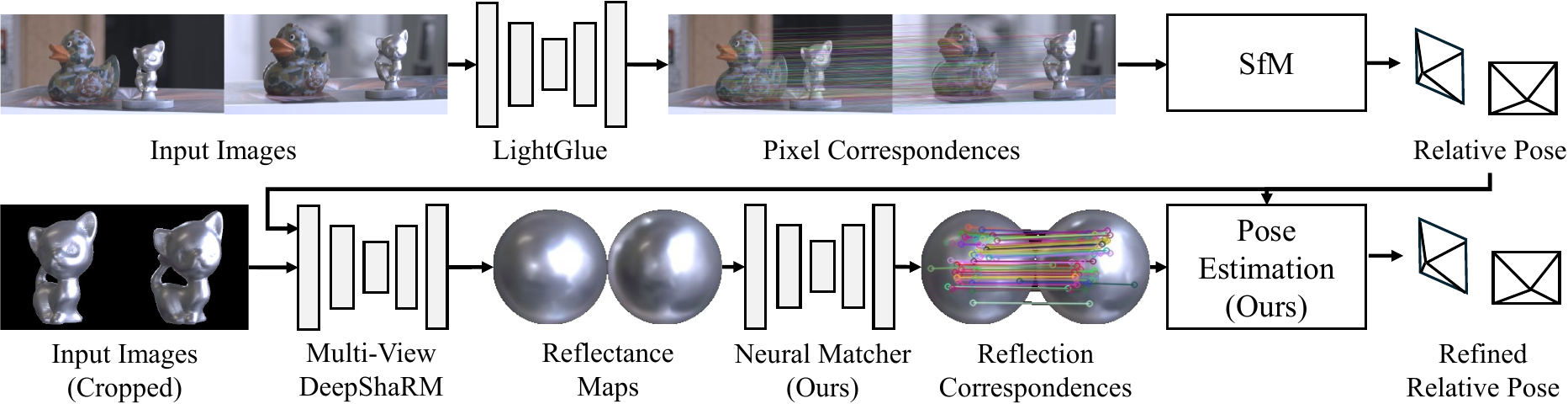}
  \caption{Overview of relative camera pose recovery from a sparse set of images capturing scenes consisting both textured and textureless, non-Lambertian objects. The obtained relative camera pose (``Refined Relative Pose'') is further improved by alternating between the multi-view DeepShaRM and our pose estimation method.}
  \label{fig:estimation-from-scene}
\end{figure}

\Cref{fig:estimation-from-scene} depicts the overall framework for relative camera pose estimation from a sparse set of images capturing scenes consisting both textured and textureless, non-Lambertian objects. We first detect pixel correspondences directly from images using an existing method~\cite{lindenberger23lightglue} and recover the relative camera pose from them by using a conventional structure-from-motion method for a perspective model. We then recover reflectance maps , normal maps, and surface geoemtry from the (initial) estimate of the relative camera pose and cropped images that capture the textureless, non-Lambertian object using DeepShaRM~\cite{yamashita2023deepsharm}. We detect reflection correspondences on the recovered reflectance maps and leverage them to obtain a refined estimate of the relative rotation and a corresponding relative translation vector.

In this setting, to avoid relying on conventional correspondences on the textureless, non-Lambertian surfaces, we do not detect 3D correspondences and do not recover the combined transformation $\mathrm{G}_{21}$. We can, however, still apply the second step of our relative rotation estimation algorithm (\cref{sec:algorithm_details}) by assuming that $\mathrm{G}_{21}$ is identical to the relative rotation $\mathrm{R}_{21}$ estimated by the SfM method. This is reasonable because multi-view consolidation of DeepShaRM ensures that discrepancies between surface normals estimated for the first view and those for the second view (\ie, the combined transformation) can be approximated by the input relative rotation. 

This enables us to apply our method to textureless, non-Lambertian objects whose pixel and 3D correspondences are ambiguous.
Note that, even in this setting, if we could detect 3D correspondences on the textureless, non-Lambertian surface, they can be beneficial especially for avoiding solutions that invert the surface geometry.
We hope to exploit pixel and 3D correspondences detected from different objects simultaneously in our future work.

Similar to Section~3.6 in the main text, we further improve the camera pose estimate by alternating between the multi-view reflectance map recovery by DeepShaRM and our camera pose estimation method.

\subsubsection{Implementation of Structure-from-Motion} We found that off-the-shelf structure-from-motion methods for perspective projection (\eg, an implementation in \cite{opencv_library}) often completely fail when cameras are almost orthographic. For a fair comparison, we improved the accuracy of the SfM baseline method by leveraging the orthographic assumption. That is, we first estimate the two Euler angles $(\phi, \theta)$ based on the orthographic model. We find a solution that maximizes the number of inlier correspondences whose reprojection errors (\cref{eq:pixel_reproj_error}) are lower than 5 pixels. We then estimate the other camera parameters by minimizing reprojection errors based on the perspective model.

\section{Dataset Details}

\begin{figure}[t]
  \centering
  \includegraphics[keepaspectratio, width=\linewidth]{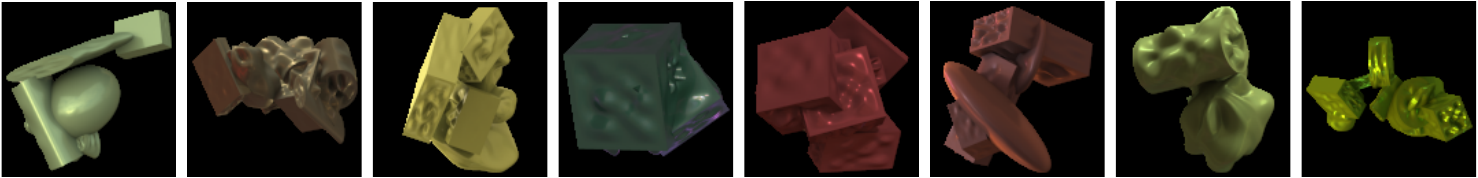}
  \caption{Example synthetic training images from the training set of the nLMVS-Synth dataset~\cite{yamashita2023nlmvs}}
  \label{fig:training-images}
\end{figure}

\Cref{fig:training-images} shows example synthetic training images from the training set of the nLMVS-Synth dataset~\cite{yamashita2023nlmvs} which we use for training the deep networks in our method. As explained in the main text, The training shapes are composed of primitive shapes (ellipsoids, cubes, cylinders) augmented with random height fields~\cite{xu2018relighting}. 94 materials and 2685 environmental maps from existing databases~\cite{matusik2003brdf,gardner2017indoor,hdrihaven} are used for rendering. 

\begin{figure*}[t]
    \centering
    \subfloat[][Illuminations]{
        \includegraphics[keepaspectratio, width=0.305\linewidth]{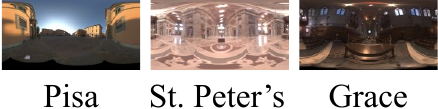}
        \label{fig:synth_test_envs}
    }
    \subfloat[][BRDFs]{
        \includegraphics[keepaspectratio, width=0.305\linewidth]{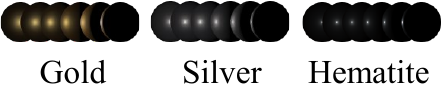}
        \label{fig:synth_test_brdfs}
    }
    \subfloat[][Shapes]{
        \includegraphics[keepaspectratio, width=0.305\linewidth]{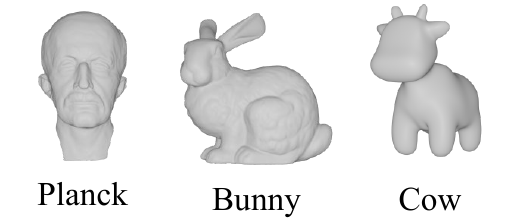}
        \label{fig:synth_test_shapes}
    }
    \caption{
        Synthetic test images in the nLMVS-Synth dataset~\cite{yamashita2023nlmvs} and some original ones are rendered using unseen illumination maps~\cite{lightprobe,hiresolightprobe}, unseen BRDFs~\cite{matusik2003brdf}, and unseen shapes~\cite{stanford3dscan,suggestivecontour,crane2013robust}.}
   
    \label{fig:synth-test-assets}
\end{figure*}

\Cref{fig:synth-test-assets} shows unseen illumination maps, unseen BRDFs, and unseen shapes used for rendering the synthetic test images in the nLMVS-Synth dataset~\cite{yamashita2023nlmvs} and some rendered by ourselves. We excluded images in the nLMVS-Synth that are rendered from less specular BRDFs (\eg, ``Rubber'') or monotonous illumination maps (\eg, ``Uffizi'') as appearance caused by them are almost independent of surface normals and extremely challenging to both our method and existing methods. Also, we could use only images of 2 shapes (``Bunny'' and ``Planck'') mainly because conventional correspondences on the other shapes (\eg, ``Sphere'') are ambiguous. Instead, we rendered images of a cow model~\cite{crane2013robust} by ourselves.

\begin{figure}[t]
  \centering
  \includegraphics[keepaspectratio, width=0.3\linewidth]{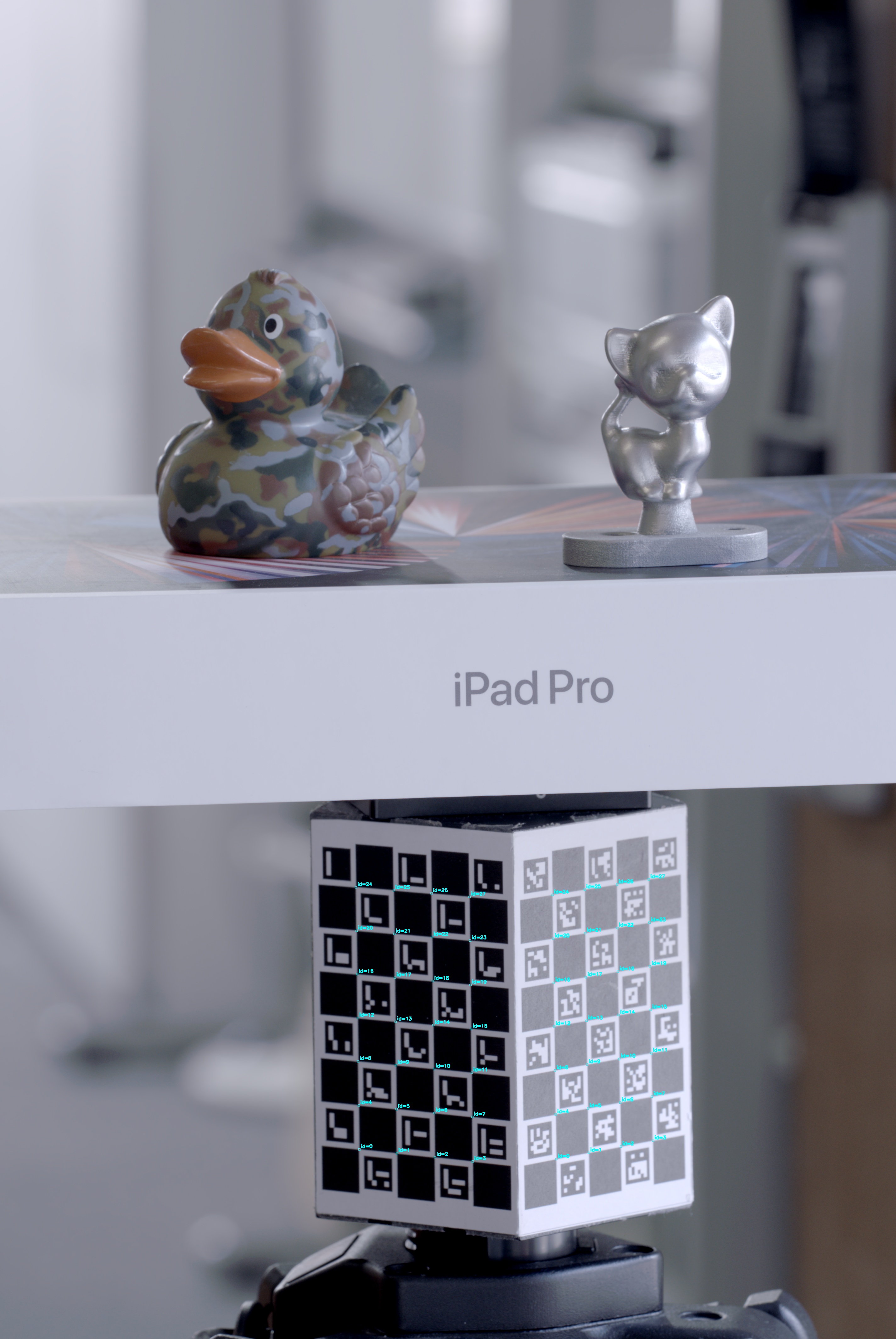}
  \caption{We use an object with ChArUco patterns for calibration of real images captured by ourselves.}
  \label{fig:real-image-example}
\end{figure}
\Cref{fig:real-image-example} shows how we calibrate the ground truth camera poses for real images captured by ourselves. We used an object with ChArUco patterns to detect 3D-2D point correspondences, and calibrated the camera parameters by using an off-the-shelf solver~\cite{opencv_library} for the Perspective-n-Point (PnP) pose computation problem.

\end{document}

%% file: vincents_commands.tex
\usepackage{dsfont}
\usepackage{etoolbox}
\usepackage{color}

\newif\ifshowedits

\newcommand{\addeditor}[3]{%
  \definecolor{#1color}{rgb}{#3}
  \expandafter\newcommand\csname #1\endcsname[1]{%
  \ifshowedits
    {\color{#1color} ##1}%
  \else
    {##1}%
  \fi
  }%
  \expandafter\newcommand\csname #1rmk\endcsname[1]{%
  \ifshowedits
    {\color{#1color} {\bf [#2: ##1]}}
  \fi
  }%
  \expandafter\newcommand\csname #1rpl\endcsname[2]{%
  \ifshowedits
    {\color{#1color} ##1 \sout{##2}}
  \else
    {##1}
  \fi
  }%
}

\newcommand{\createtextvar}[1]{
  \expandafter\newcommand\csname #1\endcsname{%
  {\text{#1}}
}%
}
\newcommand{\textvars}[1]{\forcsvlist{\createtextvar}{#1}}

\newcommand{\mycomment}[1]{}

\newcommand{\bn}{{\bf n}}

\newcommand{\bF}{{\bf F}}

\newcommand{\bN}{{\bf N}}

\DeclareMathOperator*{\argmin}{arg\,min}

\newcommand{\vcomment}[1]{}